%% file: main.tex
\documentclass[journal]{IEEEtran}
\usepackage{epsfig}
\usepackage{graphicx}
\usepackage{amsmath}
\usepackage{amssymb}

\usepackage{xcolor}
\usepackage{soul}
\usepackage[pagebackref=true,breaklinks=true,letterpaper=true,colorlinks,bookmarks=false,citecolor=cyan]{hyperref}

\usepackage{gensymb,graphics,subfigure,inputenc}
\usepackage[linesnumbered,algo2e,boxed]{algorithm2e}
\graphicspath{{Figure/}}
\usepackage[figuresright]{rotating}
\usepackage{amsmath,amssymb,textcomp,subfigure,multirow,upgreek}
\usepackage{booktabs}
\usepackage{color,mdwlist}

\usepackage{ragged2e}

\usepackage{graphicx,fontawesome}
\graphicspath{{Figures/}}

\usepackage{caption}
\usepackage{enumitem}
\setitemize{noitemsep,topsep=0pt,parsep=0pt,partopsep=0pt}

\ifCLASSINFOpdf
\else
\fi
\begin{document}
	

\title{AttentionGAN: Unpaired Image-to-Image Translation using Attention-Guided Generative Adversarial Networks}
%
%
%

\author{Hao Tang,
    Hong Liu,
	Dan Xu,
	Philip H.S. Torr,  and 
	Nicu Sebe
\thanks{Hao Tang is with the Computer Vision Lab, ETH Zurich, Switzerland.
\par Hong Liu is with the Shenzhen Graduate School, Peking University, China.
\par Dan Xu is with the Department of Computer Science and Engineering, Hong Kong University of Sciences and Technology (HKUST), Hong Kong.
\par Philip H.S. Torr is with the Department of Engineering Science, University of Oxford, United Kingdom. 
\par Nicu Sebe is with the Department of Information Engineering and Computer Science (DISI), University of Trento, Italy. 
\par Corresponding authors: Hao Tang and Hong Liu.}
}

%
%

\markboth{IEEE Transactions on Neural Networks and Learning Systems}%
{Shell \MakeLowercase{\textit{et al.}}: Bare Demo of IEEEtran.cls for IEEE Journals}
%



\maketitle



\input{0abstract}


\input{1introduction}
\input{2relatedwork}
\input{3method}
\input{4experiments}
\input{5conclusions}

\footnotesize
\bibliographystyle{IEEEtran}
\bibliography{ref}

\begin{IEEEbiography}[{\includegraphics[width=1in,height=1.25in,clip,keepaspectratio]{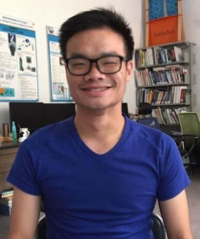}}]{Hao Tang}
is currently a Postdoctoral with Computer Vision Lab, ETH Zurich, Switzerland. 
He received the master’s degree from the School of Electronics and Computer Engineering, Peking University, China and the Ph.D. degree from Multimedia and Human Understanding Group, University of Trento, Italy.
He was a visiting scholar in the Department of Engineering Science at the University of Oxford. His research interests are deep learning, machine learning, and their applications to computer vision.
\end{IEEEbiography}
\vspace{-1cm}

\begin{IEEEbiography}[{\includegraphics[width=1in,height=1.25in,clip,keepaspectratio]{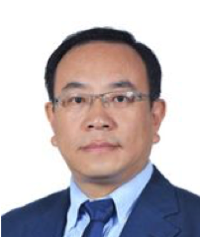}}]{Hong Liu}
received the Ph.D. degree in mechanical electronics and automation in 1996, and serves as a Full Professor in the School of EE\&CS, Peking University (PKU), China. Prof. Liu has been selected as Chinese Innovation Leading Talent supported by National High-level Talents Special Support Plan since 2013. He is also the Director of Open Lab on Human Robot Interaction, PKU. His research fields include computer vision and robotics, image processing, and pattern recognition. 
\end{IEEEbiography}
\vspace{-1cm}

\begin{IEEEbiography}[{\includegraphics[width=1in,height=1.25in,clip,keepaspectratio]{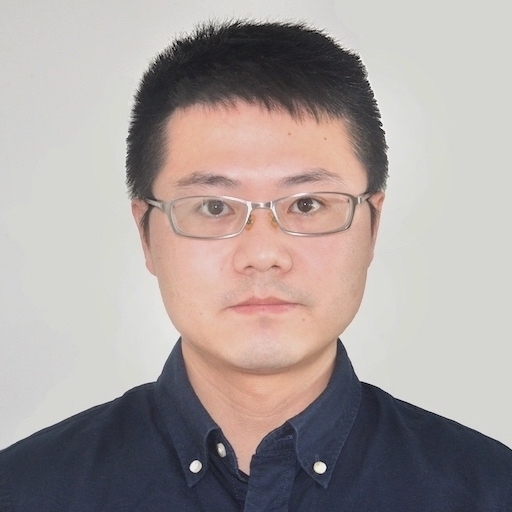}}]{Dan Xu}
is an Assistant Professor in the Department of Computer Science and Engineering at
HKUST. He was a Postdoctoral Research Fellow in VGG at the University of Oxford. He was a Ph.D. in the Department of Computer Science at the University of Trento. He was also a research assistant of MM Lab at the Chinese University of
Hong Kong. 
He served as Senior Programme Committee or Area Chair of AAAI 2022, ACM MM 2021, ACM MM 2020, ICPR 2020 and WACV 2021.
\end{IEEEbiography}

\vspace{-1cm}
\begin{IEEEbiography}[{\includegraphics[width=1in,height=1.25in,clip,keepaspectratio]{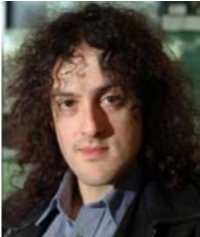}}]{Philip H.S. Torr} is Professor at the University of Oxford, UK. After working for another three
years at Oxford, he worked for six years for
	Microsoft Research, first in Redmond, then in
	Cambridge, founding the vision side of the Machine Learning and Perception Group. He has won
	awards from top vision conferences, including
	ICCV, CVPR, ECCV, NIPS and BMVC. He is a
	senior member of the IEEE and a Royal Society
	Wolfson Research Merit Award holder.
\end{IEEEbiography}

\vspace{-1cm}
\begin{IEEEbiography}[{\includegraphics[width=1in,height=1.25in,clip,keepaspectratio]{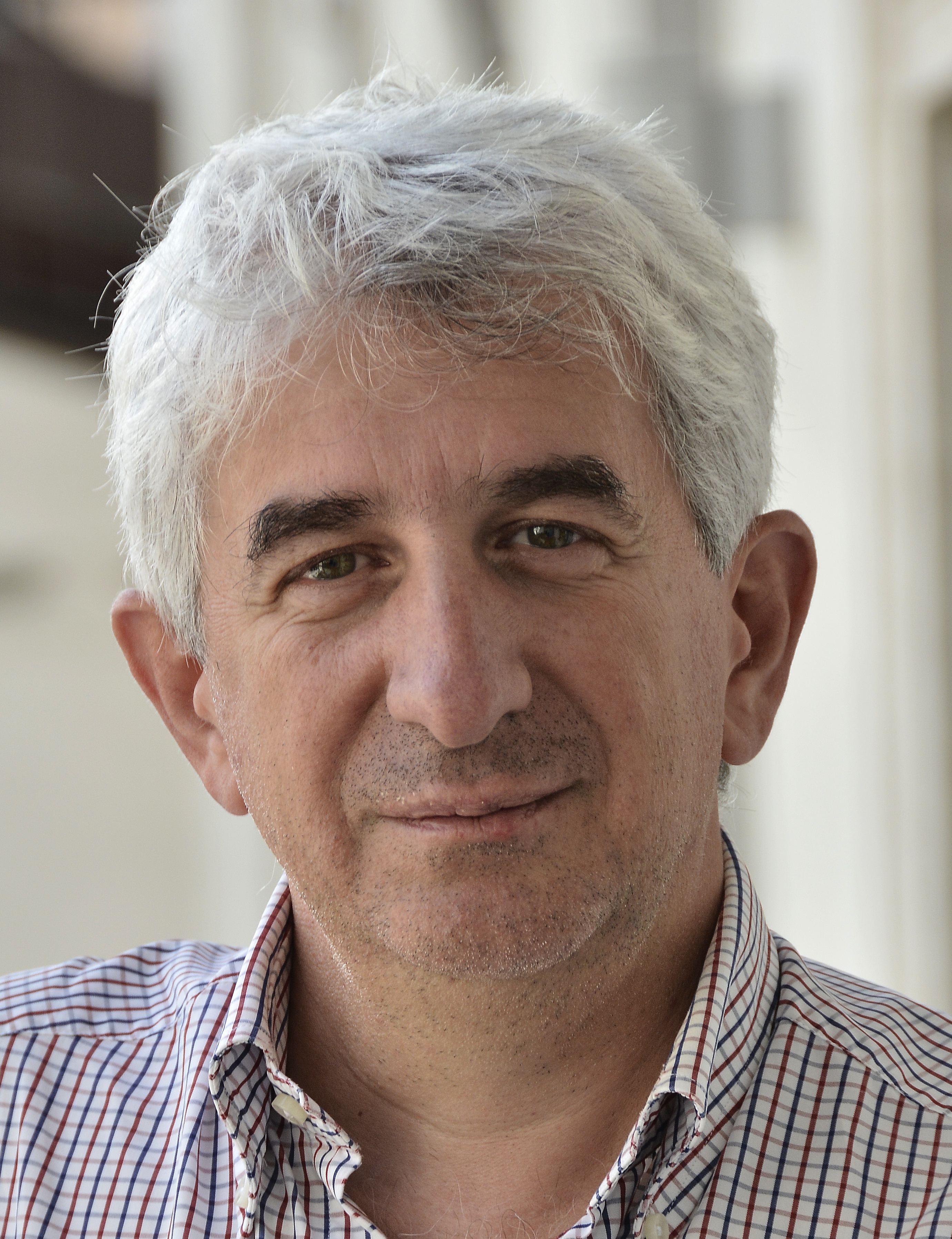}}]{Nicu Sebe} is Professor in the University of Trento, Italy, where he is leading the research in the areas of multimedia analysis and human behavior understanding. He was the General Co-Chair of the IEEE FG 2008 and ACM Multimedia 2013.  He was a program chair of ACM Multimedia 2011 and 2007, ECCV 2016, ICCV 2017 and ICPR 2020.  He is a general chair of ACM Multimedia 2022 and a program chair of ECCV 2024. He is a fellow of IAPR.
\end{IEEEbiography}

\end{document}

%% file: 0abstract.tex
\begin{abstract}
State-of-the-art methods in image-to-image translation are capable of learning a mapping from a source domain to a target domain with unpaired image data.
Though the existing methods have achieved promising results, they still produce visual artifacts, being able to translate low-level information but not high-level semantics of input images.
One possible reason is that generators do not have the ability to perceive the most discriminative parts between the source and target domains, thus making the generated images low quality.
In this paper, we propose a new Attention-Guided Generative Adversarial Networks (AttentionGAN) for the unpaired image-to-image translation task.	
AttentionGAN can identify the most discriminative foreground objects and minimize the change of the background.
The attention-guided generators in AttentionGAN are able to produce attention masks, and then fuse the generation output with the attention masks to obtain high-quality target images. 
Accordingly, we also design a novel attention-guided discriminator which only considers attended regions.
Extensive experiments are conducted on several generative tasks with eight public datasets, demonstrating that the proposed method is effective to generate sharper and more realistic images compared with existing competitive models. The code is available at~\url{https://github.com/Ha0Tang/AttentionGAN}.
\end{abstract}

%% file: 1introduction.tex
\section{Introduction}

Recently, Generative Adversarial Networks (GANs)~\cite{goodfellow2014generative} in various fields such as computer vision and image processing have produced powerful translation systems with supervised settings such as Pix2pix~\cite{isola2016image}, where paired training images are required.
However, paired data are usually difficult or expensive to be obtained. 
The input-output pairs for tasks such as artistic stylization could be even more difficult to be acquired since the desired output is quite complex, typically requiring artistic authoring. 
To tackle this problem, CycleGAN~\cite{zhu2017unpaired}, DualGAN~\cite{yi2017dualgan} and DiscoGAN~\cite{kim2017learning} provide a new insight, in which the GAN models can learn the mapping from a source domain to a target one with unpaired image data.

\begin{figure}[!t] \footnotesize
	\centering
	\includegraphics[width=0.85\linewidth]{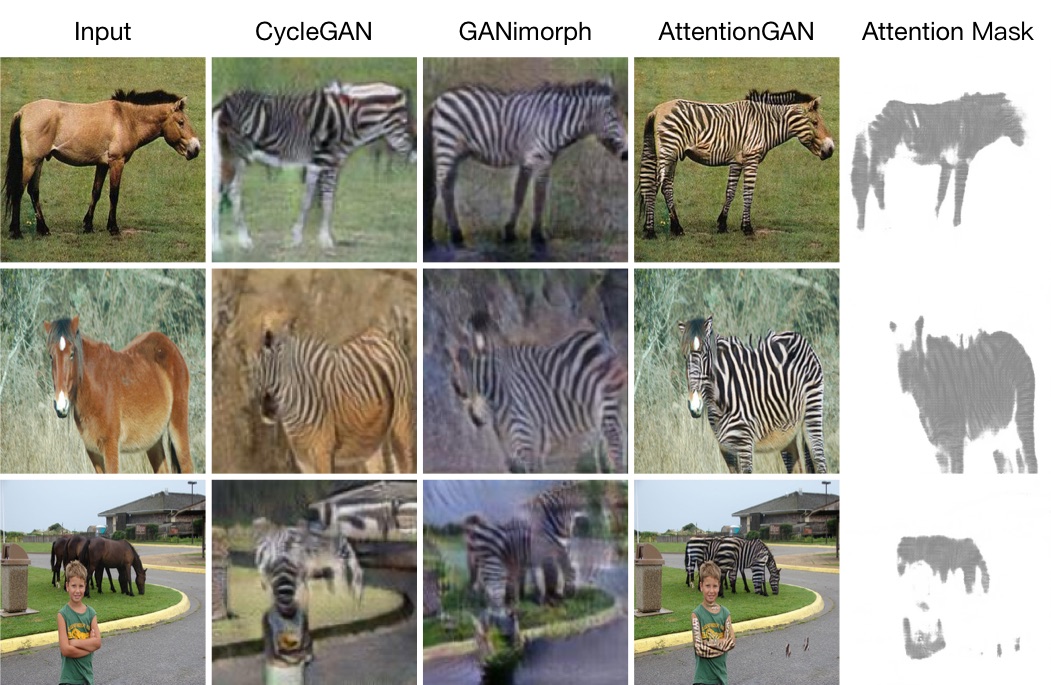}
	\caption{Comparison with existing unpaired image-to-image translation methods (e.g., CycleGAN~\cite{zhu2017unpaired} and GANimorph~\cite{gokaslan2018improving}) with an example of horse to zebra translation. We are interested in transforming horses to zebras. In this case we should be agnostic to the background. However methods such as CycleGAN and GANimorph will transform the background in a nonsensical way, in contrast to our attention-based method (the generated attention masks are shown in the last column).
	}
	\label{fig:task}
	\vspace{-0.4cm}
\end{figure}

\begin{figure*}[!t] \footnotesize
	\centering
	\includegraphics[width=0.85\linewidth]{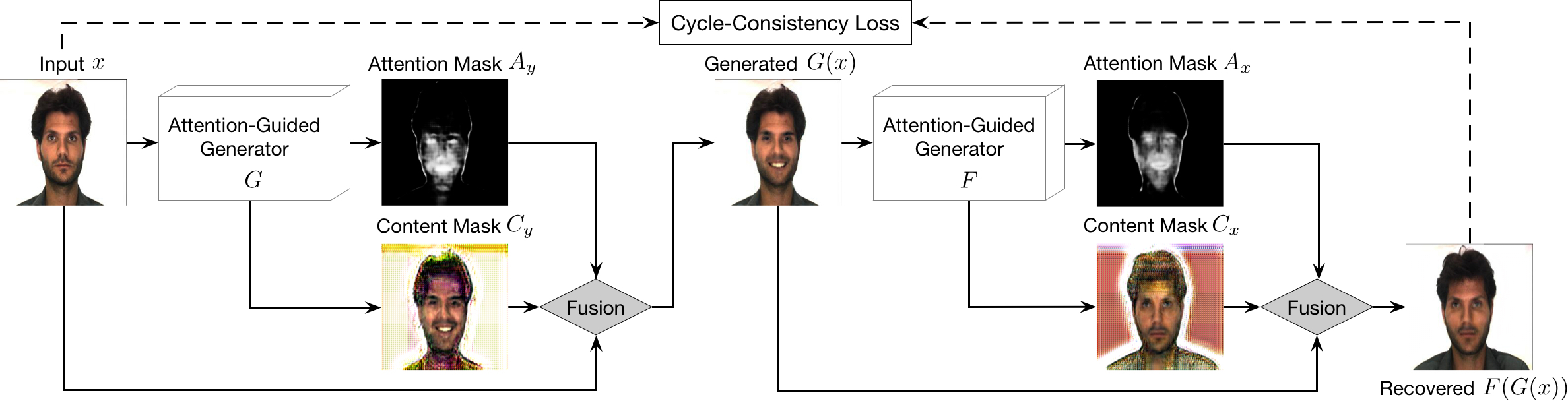}
	\caption{Framework of the proposed attention-guided generation scheme I, which consists of two attention-guided generators $G$ and $F$. We show one mapping in this figure, i.e., $x {\rightarrow} G(x) {\rightarrow} F(G(x)) {\approx} x$. We also have the other mapping, i.e., $y {\rightarrow} F(y) {\rightarrow}  G(F(y)) {\approx} y$. The attention-guided generators have a built-in attention module, which can perceive the most discriminative content between the source and target domains. We then fuse the input image, the content mask, and the attention mask to synthesize the final result.}
	\label{fig:excyclegan}
	\vspace{-0.4cm}
\end{figure*}

Despite these efforts, unpaired image-to-image translation remains a challenging problem. Most existing models change unwanted parts in the translation and can also be easily affected by background changes (see Fig.~\ref{fig:task}). 
In order to address these limitations, Liang et al.~propose ContrastGAN~\cite{liang2017generative}, which uses object-mask annotations provided by the dataset to guide the generation,  first cropping the unwanted parts in the image based on the masks, and then pasting them back after the translation.
While the generated results are reasonable, it is hard to collect training data with object-mask annotations. 
Another option is to train an extra model to detect the object masks and then employ them for the mask-guided generation~\cite{chen2018attention,kastaniotis2018attention}.
In this case, we need to significantly increase the network capacity, which consequently raises the training complexity in both time and space.

To overcome the aforementioned issues, we propose a novel Attention-Guided Generative Adversarial Networks (AttentionGAN) for the unpaired image-to-image translation task.
Fig.~\ref{fig:task} shows a comparison with exiting image-to-image translation methods using a horse to zebra translation example. 
The most important advantage of AttentionGAN is that the proposed generators can focus on the foreground of the target domain and preserve the background of the source domain effectively. 
Specifically, the proposed generator learns both foreground and background attentions. It uses the foreground attention to select from the generated output for the foreground regions, while uses the background attention to maintain the background information from the input image. 
In this way, the proposed AttentionGAN can focus on the most discriminative foreground and ignore the unwanted background.
We observe that AttentionGAN achieves significantly better results than both GANimorph~\cite{gokaslan2018improving} and CycleGAN~\cite{zhu2017unpaired}. 
As shown in Fig.~\ref{fig:task}, AttentionGAN not only produces clearer results, but also successfully maintains the little boy in the background and only performs the translation for the horse behind it.
However, the existing holistic image-to-image translation approaches are generally interfered by irrelevant background content, thus hallucinating texture patterns of the target objects.

We propose two different attention-guided generation schemes for the proposed AttentionGAN.
The framework of the proposed scheme I is shown in Fig.~\ref{fig:excyclegan}.
The proposed generators are equipped with a built-in attention module, which can disentangle the discriminative semantic objects from the unwanted parts via producing an attention mask and a content mask.
Then we fuse the attention and the content masks to obtain the final generation. 
Moreover, we design two novel attention-guided discriminators which aim to consider only the attended foreground regions.
The proposed attention-guided generators and discriminators are trained in an end-to-end fashion.
The proposed attention-guided generation scheme I can achieve promising results on the facial expression translation (see Fig.~\ref{fig:scheme12}), where the change between the source domain and the target domain is relatively minor.
However, it performs unsatisfactorily on more challenging scenarios in which a more complex semantic translation is required such as horse to zebra translation as shown in Fig.~\ref{fig:task}. 
To tackle this issue, we further propose a more advanced attention-guided generation scheme~II, as depicted in Fig.~\ref{fig:v2}.
The improvement upon the scheme~I is mainly three-fold: 
First, in scheme I the attention and the content masks are generated with the same network. 
To have a more powerful generation of them, we employ two separate sub-networks in scheme~II.
Second, in scheme I we only generate the foreground attention mask to focus on the most discriminative semantic content.
However, to better learn the foreground and preserve the background simultaneously, we produce both foreground and background attention masks in scheme II.
Third, as the foreground generation is more complex, instead of learning a single content mask in the scheme I, we learn a set of several intermediate content masks, and correspondingly we also learn the same number of foreground attention masks. The generation of multiple intermediate content masks is beneficial for the network to learn a more rich generation space. 
The intermediate content masks are then fused with the foreground attention masks to produce the final content masks.
Extensive experiments on several challenging public benchmarks demonstrate that the proposed scheme II can produce higher-quality target images compared with existing state-of-the-art methods.

The contribution of this paper is summarized as follows:
\begin{itemize} [leftmargin=*]
	\item We propose a new Attention-Guided Generative Adversarial Network (AttentionGAN) for the unpaired image-to-image translation task. 
	This framework stabilizes the GANs’ training and thus improves the quality of generated images through jointly approximating attention and content masks with several losses and optimization methods.
	\item We design two novel attention-guided generation schemes for the proposed framework, to better perceive and generate the most discriminative foreground parts and simultaneously preserve well the unfocused objects and background. 
	Moreover, the proposed attention-guided generator and discriminator can be flexibly applied in other GANs to improve the multi-domain image-to-image translation task, which we believe would also be beneficial to other related research.
	\item We conduce extensive experiments on eight popular datasets and experimental results show that the proposed AttentionGAN can generate photo-realistic images with more clear details compared with existing methods. We also established new state-of-the-art results on these datasets.
\end{itemize}

Part of the material presented here appeared in \cite{tang2019attention}. The current paper extends \cite{tang2019attention} in several ways.
1) We propose a more advanced attention-guided generation scheme, i.e., the scheme II, which is a more robust and general framework for both unpaired image-to-image translation and multi-domain image translation tasks. 
2) We present an in-depth description of the proposed approach, providing all the architectural and implementation details of the method, with special emphasis on guaranteeing the reproducibility of the experiments. The source code is also available online.
3) We extend the experimental evaluation provided in \cite{tang2019attention} in several directions. 
First, we conduct extensive experiments on eight popular datasets, demonstrating the wide application scope of the proposed AttentionGAN.
Second, we conduct exhaustive ablation studies to evaluate each component
of the proposed method.
Third, we investigate the influence of hyper-parameters on generation performance.
Fourth, we provide the visualization of the learned attention masks to better interpret our model. 
Fifth, we have also included more state-of-the-art baselines, e.g.,  UAIT~\cite{mejjati2018unsupervised}, U-GAT-IT~\cite{kim2019ugatit}, SAT~\cite{yang2019show}, MUNIT~\cite{huang2018multimodal}, DRIT~\cite{lee2018diverse}, GANimorph~\cite{gokaslan2018improving}, TransGaGa \cite{wu2019transgaga}, DA-GAN \cite{ma2018gan}, and we observe that AttentionGAN achieves better results than those. 
Lastly, we conduct extensive experiments on the multi-domain image translation task, and experimental results show that AttentionGAN achieves much better results than existing leading methods such as StarGAN~\cite{choi2017stargan}.

%% file: 2relatedwork.tex
\section{Related Work}
\label{sec:related} 

\noindent \textbf{Generative Adversarial Networks (GANs)}~\cite{goodfellow2014generative} are powerful generative models, which have achieved impressive results on different computer vision tasks, e.g.,
image/video generation~\cite{liu2021cross,tang2021total}.
To generate meaningful images that meet user requirements, Conditional GANs (CGANs)~\cite{mirza2014conditional,tang2020unified} inject extra information to guide the image generation process, which can be discrete labels~\cite{perarnau2016invertible,choi2017stargan},
object keypoints~\cite{tang2019cycle}, human skeleton~\cite{tang2018gesturegan,tang2020bipartite,tang2020xinggan}, and semantic maps~\cite{tang2019selection,tang2020dual,tang2020local}. 

\noindent \textbf{Image-to-Image Translation} models learn a translation function using CNNs. 
Pix2pix~\cite{isola2016image} is a conditional framework using a CGAN to learn a mapping function from input to output images. 
Wang et al.~propose Pix2pixHD~\cite{wang2018pix2pixHD} for high-resolution photo-realistic image-to-image translation, which can be used for turning semantic label maps into photo-realistic images.
Similar ideas have also been applied to many other tasks, such as hand gesture generation~\cite{tang2018gesturegan}.
However, most of the tasks in the real world suffer from having few or none of the paired input-output samples available.
When paired training data is not accessible, image-to-image translation becomes an ill-posed problem.

\noindent \textbf{Unpaired Image-to-Image Translation.}
To overcome this limitation, the unpaired image-to-image translation task has been proposed.
In this task, the approaches learn the mapping function without the requirement of paired training data.
Specifically, CycleGAN~\cite{zhu2017unpaired} learns the mapping between two image domains instead of paired images.
Apart from CycleGAN, many other GAN variants \cite{kim2017learning,yi2017dualgan,benaim2017one,anoosheh2017combogan,tang2018dual,choi2017stargan,wang2018mix} have been proposed to tackle the cross-domain problem.
However, those models can be easily affected by unwanted content and cannot focus on the most discriminative semantic part of images during the translation stage.

\noindent \textbf{Attention-Guided Image-to-Image Translation.}
To fix the aforementioned limitations, several works employ an attention mechanism to help image translation.
Attention mechanisms have been successfully introduced in many applications in computer vision such as depth estimation~\cite{xu2018structured},
helping the models to focus on the relevant portion of the input.

Recent works use attention modules to attend to the region of interest for the image translation task in an unsupervised way, which can be divided into two categories.
The first category is to use extra data to provide attention.
For instance, 
Liang~et al. propose ContrastGAN~\cite{liang2017generative}, which uses the object mask annotations from each dataset as extra input data.
Moreover, Mo et al.~propose InstaGAN~\cite{mo2018instagan} that incorporates instance information 
(e.g., object segmentation masks) 
and improves multi-instance transfiguration.

The second type is to train another segmentation or attention model to generate attention maps and fit it to the system.
For example,
Chen et al.~\cite{chen2018attention} use an extra attention network to generate attention maps, so that more attention can be paid to objects of interests.
Kastaniotis et al.~present ATAGAN~\cite{kastaniotis2018attention}, which uses a teacher network to produce attention maps.
Yang et al.~\cite{yang2019show} propose to add an attention module to predict an attention map to guide the image translation process. 
Kim et al.~\cite{kim2019ugatit} propose to use an auxiliary classifier to generate attention masks.
Mejjati et al.~\cite{mejjati2018unsupervised} propose attention mechanisms that are jointly trained with the generators, discriminators, and other two attention networks.

All these methods employ extra networks or data to obtain attention masks, which increases the number of parameters, training time, and storage space of the whole system.
Moreover, we still observe unsatisfactory aspects in the generated images by these methods.
To fix both limitations,  we propose  a novel AttentionGAN, which aims to disentangle the input image into foreground and background by generating multiple foreground attention masks, a background attention mask, and multiple content masks.
Then AttentionGAN learns to attend to key parts of the image and translate the salient objects/foreground to the target domain while keeping everything else unaltered, essentially avoiding undesired artifacts or changes.
In this way, we do not need any extra models to obtain the attention masks of objects of interests.
To the best of our knowledge, this idea has not been investigated by existing attention-guided image-to-image translation methods.
Also, extensive experiments show that  AttentionGAN achieves significantly better results than the existing attention-guided generation methods.
Most importantly, our method can be applied to any GAN-based framework such as unpaired, paired, and multi-domain image-to-image translation frameworks.

%% file: 3method.tex
\section{Attention-Guided GANs}
\label{sec:method}

\begin{figure*}[!t] \footnotesize
	\centering
	\includegraphics[width=0.83\linewidth]{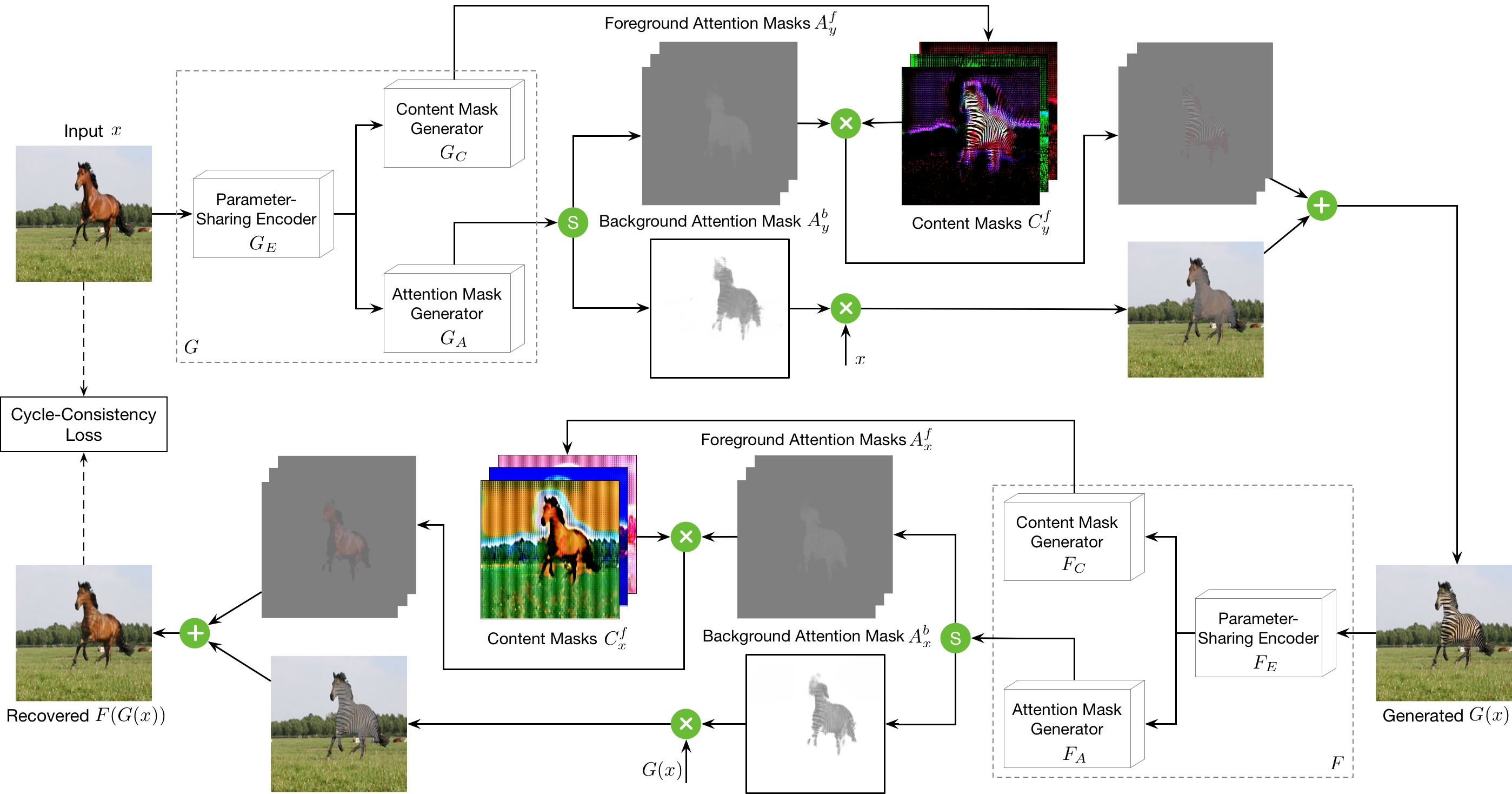}
	\caption{Framework of the proposed attention-guided generation scheme II, which consists of two attention-guided generators $G$ and $F$. 
		We show one mapping in this figure, i.e., $x {\rightarrow} G(x) {\rightarrow} F(G(x)) {\approx} x$. 
		We also have the other mapping, i.e., $y {\rightarrow} F(y) {\rightarrow}  G(F(y)) {\approx} y$. 
		Each generator such as $G$ consists of a parameter-sharing encoder $G_E$, an attention mask generator $G_A$, and a content mask generator $G_C$. 
		$G_A$ aims to produce attention masks of both foreground and background to attentively select the useful content from the corresponding content masks generated by $G_C$ and the input image $x$. The proposed model is constrained by the cycle-consistency loss and trained in an end-to-end fashion so that each generator can benefit from each other. The symbols $\oplus$, $\otimes$ and $\textcircled{s}$ denote element-wise addition, multiplication, and channel-wise Softmax, respectively.}
	\label{fig:v2}
	\vspace{-0.4cm}
\end{figure*}

\subsection{Attention-Guided Generation}
GANs~\cite{goodfellow2014generative} are composed of two competing modules: the generator $G$ and the discriminator $D$, which are iteratively trained competing against each other in the manner of two-player mini-max.
More formally, let $X$ and $Y$ denote two different image domains, $x_i{\in} X$ and $y_j{\in} Y$ denote the training images in $X$ and $Y$, respectively (for simplicity, we usually omit the subscript $i$ and~$j$).
For most current image translation models, e.g., CycleGAN~\cite{zhu2017unpaired} and DualGAN~\cite{yi2017dualgan}, they include two generators $G$ and $F$, and two corresponding adversarial discriminators $D_X$ and $D_Y$.
Generator $G$ maps $x$ from the source domain to the generated image $G(x)$ in the target domain $Y$ and tries to fool the discriminator $D_Y$, whilst $D_Y$ focuses on improving itself to be able to tell whether a sample is a generated sample or a real data sample.
Similar to generator $F$ and discriminator $D_X$.

\noindent \textbf{Attention-Guided Generation Scheme I.}
For the proposed AttentionGAN, we learn two mappings between domains $X$ and $Y$ via two generators with built-in attention mechanism, i.e., $G{:} x {\rightarrow} [A_y, C_y] {\rightarrow} G(x)$ and $F{:} y {\rightarrow} [A_x, C_x] {\rightarrow} F(y)$, where $A_x$ and $A_y$ are the attention masks of images $x$ and $y$, respectively; 
$C_x$ and $C_y$ are the content masks of images $x$ and $y$, respectively;
$G(x)$ and $F(y)$ are the generated images.
The attention masks $A_x$ and $A_y$ define a per pixel intensity specifying to which extent each pixel for the content masks $C_x$ and $C_y$ will contribute in the final rendered image.
In this way, the generator does not need to render static elements (it refers to background), and can focus exclusively on the pixels defining the domain content movements, leading to sharper and more realistic synthetic images.
After that, we fuse the input image $x$, the generated attention mask $A_y$, and the content mask $C_y$ to obtain the targeted image $G(x)$. 
In this way, we can disentangle the most discriminative semantic objects and the unwanted part of images.
Take Fig.~\ref{fig:excyclegan} as an example, the attention-guided generators focus only on those regions of the image that are responsible of generating the novel expression such as eyes and mouth, and keeping the rest of parts of the image such as hair, glasses, and clothes untouched.
The higher intensity in the attention mask means the larger contribution for changing the expression.

The input of each generator is a three-channel image, and the outputs of each generator are an attention mask and a content mask.
Specifically, the input image of generator $G$ is $x{\in} \mathbb{R}^{H\times W \times 3}$, and the outputs are the attention mask $A_y {\in} \{0,...,1\}^{H\times W}$ and content mask $C_y {\in} \mathbb{R}^{H\times W \times 3}$.
Thus, we use the following formula to produce the final image~$G(x)$,
\begin{equation}
	\begin{aligned}
		G(x) = C_y * A_y + x * (1 - A_y),
	\end{aligned}
	\label{eqn:atten}
\end{equation}  
where the attention mask $A_y$ is copied to three channels for multiplication purpose.
Intuitively, the attention mask $A_y$ enables some specific areas where domain changed to get more focus and applying it to the content mask $C_y$ can generate images with clear dynamic area and unclear static area.
The static area should be similar between the generated image and the original real image. 
Thus, we can enhance the static area in the original real image $x *(1 {-} A_y )$ and merge it to $C_y {*} A_y$ to obtain final result $C_y {*} A_y + x {*} (1 {-} A_y)$.
Similarly, the formulation for generator $F$ and input image $y$ can be expressed as $F(y) {=} C_x * A_x {+} y * (1 {-} A_x)$.

The proposed attention-guided generation scheme I performs well on the tasks where the source domain and the target domain have large overlap similarity, such as facial expression-to-expression translation.
However, we observe that it cannot generate photo-realistic images on complex tasks such as horse to zebra translation, as shown in Fig.~\ref{fig:scheme12}.
The drawbacks of the scheme I are three-fold:
(i) The attention and the content mask are generated by the same network, which could degrade the quality of the generated images.
(ii) We observe that the scheme~I only produces one attention mask to simultaneously change the foreground and preserve the background of the input images.
(iii) We see that scheme I only produces one content mask to select useful content for generating the foreground content, which means the model does not have enough ability to deal with complex tasks such as horse to zebra translation.
To solve these limitations, we further propose a more advanced attention-guided generation scheme II as shown in Fig.~\ref{fig:v2}.

\noindent \textbf{Attention-Guided Generation Scheme II.}
Scheme I adopts the same network to produce both attention and content masks, and we argue that this will degrade the generation performance.
In scheme II, the proposed generators $G$ and $F$ are composed of two sub-nets each for generating attention masks and content masks as shown in Fig.~\ref{fig:v2}.
For instance, $G$ is comprised of a parameter-sharing encoder $G_E$, an attention mask generator $G_A$, and a content mask generator $G_C$.
$G_E$ aims at extracting both low-level and high-level deep feature representations.
$G_C$ targets to produce multiple intermediate content masks.
$G_A$ tries to generate multiple attention masks. 
By the way, both attention mask generation and content mask generation have their own network parameters and will not interfere with each other.

To fix the limitation (ii) of scheme I, in scheme II, the attention mask generator $G_A$ targets to generate both $n{-}1$ foreground attention masks $\{A_y^f\}_{f=1}^{n-1}$ and one background attention mask~$A_y^b$, as shown in Fig.~\ref{fig:v2}.
By doing so, the proposed network can simultaneously learn the novel foreground and preserve the background of input images.
The key point success of the proposed scheme II are the generation of both foreground and background attention masks, which allow the model to modify the foreground and simultaneously preserve the background of input images.
This is exactly the goal that the unpaired image-to-image translation tasks aim to optimize.

Moreover, we observe that in some generation tasks such as horse to zebra translation, the foreground generation is very difficult if we only produce one content mask as done in scheme~I.
To fix this limitation, we use the content mask generator $G_C$ to produce $n{-}1$ content masks $\{C_y^f\}_{f=1}^{n-1}$, as shown in Fig.~\ref{fig:v2}.
Then take the input image $x$ into consideration, we obtain $n$ intermediate content masks.
In this way, a three-channel generation space can be enlarged to a $3n$-channel generation space, which is more suitable for learning a good mapping for complex image translation.

Specifically, the feature map $m$ extracted from $G_E$ is first fed into the generator $G_C$ to generate $n{-}1$ content masks $\{C_y^f\}_{f=1}^{n-1}$, followed by a $\rm{Tanh(\cdot)}$ activation function.
This process can be expressed as,
\begin{equation}
\begin{aligned}
C_y^f= {}& {\rm Tanh}(m W_C^f +b_C^f),  & \quad   {\rm for}~f  =  1, \cdots, n-1
\end{aligned}
\end{equation}
where a convolution operation is performed with $n{-}1$ convolutional filters $\{W_C^f, b_C^f\}_{f=1}^{n-1}$.
Thus, the $n{-}1$ content masks and the input image $x$ can be regarded as the candidate image pool.

Meanwhile, the feature map $m$ is fed into a group of filters $\{W_A^f, b_A^f\}_{f=1}^{n}$ to generate the corresponding $n$ attention masks,
\begin{equation}
\begin{aligned}
A_y^f =  {\rm Softmax}(m W_A^f +b_A^f),  & \quad   {\rm for}~f =  1, \cdots, n
\end{aligned}
\end{equation}
where ${\rm Softmax(\cdot)}$ is a channel-wise Softmax function used for the normalization. We then split $\{A_y^f\}_{f=1}^n$ into $n{-}1$ foreground attention masks $\{A_y^f\}_{f=1}^{n-1}$ and one background attention mask~$A_y^b$ along the channel dimension.
Note that the generated one background attention mask and $n{-}1$ foreground attention masks are complementary, but the generated $n{-}1$ foreground attention masks are not complementary to each other.
By doing so, the proposed model has a more flexible ability to learn and translate the foreground content.

Finally, the attention masks are multiplied by the corresponding content masks to obtain the final target result as shown in Fig.~\ref{fig:v2}. 
Formally, this is written as:
\begin{equation}
	\begin{aligned}
		G(x) = \sum_{f=1}^{n-1} (C_y^f * A_y^f) + x * A_y^b.
	\end{aligned}
	\label{eqn:atten3}
\end{equation}
In this way, we can preserve the background of the input image $x$, i.e., $x {*} A_y^b$, and simultaneously generate the novel foreground content for the input image, i.e., $\sum_{f=1}^{n-1}(C_y^f{*}A_y^f)$.
Then, we merge the generate foreground $\sum_{f=1}^{n-1}(C_y^f{*}A_y^f)$ to the background $x {*} A_y^b$ for obtaining the final result $G(x)$.
The formulation of generator $F$ and input image $y$ can be expressed as $F(y) {=} \sum_{f=1}^{n-1} (C_x^f * A_x^f) + y * A_x^b$, where $n$ attention masks $[\{A_x^f\}_{f=1}^{n-1}, A_x^b]$ are also produced by a channel-wise Softmax activation function for the normalization.

\subsection{Attention-Guided Cycle}
To further regularize the mappings, CycleGAN~\cite{zhu2017unpaired} adopts two cycles in the generation process.
The motivation of cycle-consistency is that if we translate from one domain to the other and back again, we should arrive at where we started.
Specifically, for each image $x$ in domain $X$, the image translation cycle should be able to bring $x$ back to the original one, i.e., $x{\rightarrow}G(x){\rightarrow}F(G(x)){\approx}x$.
Similarly, for the image $y$, we have another cycle, i.e., $y{\rightarrow}F(y){\rightarrow}G(F(y)){\approx}y$.
These behaviors can be achieved by using a cycle-consistency loss:
\begin{equation}
\begin{aligned}
\mathcal{L}_{cycle}(G, F) = & \mathbb{E}_{x\sim{p_{\rm data}}(x)}[\Arrowvert F(G(x))-x\Arrowvert_1]\\
+  &  \mathbb{E}_{y\sim{p_{\rm data}}(y)}[\Arrowvert G(F(y))-y\Arrowvert_1],
\end{aligned}
\label{equ:cycleganloss2}
\end{equation}
where the reconstructed image $F(G(x))$ is closely matched to the input image $x$, and is similar to the generated image $G(F(y))$ and the input image~$y$.
This could lead to generators to further reduce the space of possible mappings.

We also adopt the cycle-consistency loss in the proposed attention-guided generation scheme I and II.
However, we have modified it for the proposed models. 

\noindent \textbf{Attention-Guided Generation Cycle I.}
For the proposed attention-guided generation scheme I, we should push back the generated image $G(x)$ in Eq.~\eqref{eqn:atten} to the original domain.
Thus, we introduce another generator $F$, which has a similar structure to the generator $G$ (see Fig.~\ref{fig:excyclegan}).
Different from CycleGAN, the proposed $F$ tries to generate a content mask $C_x$ and an attention mask $A_x$.
Therefore, we fuse both masks and the generated image $G(x)$ to reconstruct the original input image $x$.
This process can be formulated as,
\begin{equation}
\begin{aligned}
F(G(x)) = C_x * A_x + G(x) * (1 - A_x),
\end{aligned}
\label{eqn:atten1}
\end{equation}  
where the reconstructed image $F(G(x))$ should be very close to the original one $x$.
For image $y$, we can reconstruct it by using $G(F(y)) {=} C_y * A_y {+} F(y) * (1 {-} A_y)$, and the recovered image $G(F(y))$ should be very close to $y$.

\noindent \textbf{Attention-Guided Generation Cycle II.}
For the proposed attention-guided generation scheme II, after generating the result $G(x)$ by the generator $G$ in Eq.~\eqref{eqn:atten3}, we should push $G(x)$ back to the original domain to reduce the space of possible mappings.
Thus, we have another generator $F$, which is very different from the one in scheme I.
$F$ has a similar structure to the generator $G$ and also consists of three sub-nets, i.e., a parameter-sharing encoder $F_E$, an attention mask generator $F_A$, and a content mask generator $F_C$ (see Fig.~\ref{fig:v2}).
$F_C$ tries to generate $n{-}1$ content masks (i.e., $\{C_x^f\}_{f=1}^{n-1}$) and $F_A$ tries to generate $n$ attention masks of both background and foreground (i.e., $A_x^b$ and  $\{A_x^f\}_{f=1}^{n-1}$).
Then we fuse these masks and the generated image $G(x)$ to reconstruct the original input image $x$.
This process can be formulated as,
\begin{equation}
\begin{aligned}
F(G(x)) = \sum_{f=1}^{n-1} (C_x^f * A_x^f) + G(x) * A_x^b,
\end{aligned}
\label{eqn:atten2}
\end{equation}
where the reconstructed image $F(G(x))$ should be very close to the original one $x$.
For image $y$, we have the cycle $G(F(y)) {=} \sum_{f=1}^{n-1} (C_y^f * A_y^f) + F(y) * A_y^b$, and the recovered image $G(F(y))$ should be very close to $y$.

\subsection{Attention-Guided Discriminator}
Eq.~\eqref{eqn:atten} constrains the generators to act only on the attended regions. 
However, the discriminators currently consider the whole image.
More specifically, the vanilla discriminator $D_Y$ takes the generated image $G(x)$ or the real image $y$ as input and tries to distinguish them, this adversarial loss  can be formulated as follows:
\begin{equation}
\begin{aligned}
\mathcal{L}_{GAN}(G, D_Y) = &  \mathbb{E}_{y\sim{p_{\rm data}}(y)}\left[ \log D_Y(y)\right] \\
+  & \mathbb{E}_{x\sim{p_{\rm data}}(x)}[\log (1 - D_Y(G(x)))],
\end{aligned}
\label{equ:basicgan}
\end{equation}
where $G$ tries to minimize the adversarial loss $\mathcal{L}_{GAN}(G, D_Y)$ while $D_Y$ tries to maximize it.
The target of $G$ is to generate an image $G(x)$ that looks similar to the images from domain $Y$, while $D_Y$ aims to distinguish between the generated image $G(x)$ and the real one $y$.
A similar adversarial loss of the generator $F$ and its discriminator $D_X$ is defined as,
\begin{equation}
\begin{aligned}
\mathcal{L}_{GAN}(F, D_X) = &  \mathbb{E}_{x\sim{p_{\rm data}}(x)}[\log D_X(x)] \\
+ & \mathbb{E}_{y\sim{p_{\rm data}}(y)}[\log (1 {-} D_X(F(y)))],
\end{aligned}
\label{equ:basicgan2}
\end{equation}
where $D_X$ tries to distinguish between the generated image $F(y)$ and the real one~$x$.
  
To add an attention mechanism to the discriminator, we propose two attention-guided discriminators.
The attention-guided discriminator is structurally the same as the vanilla discriminator but it also takes the attention mask as input.
The attention-guided discriminator $D_{YA}$, tries to distinguish between the fake image pairs $[A_y, G(x)]$ and the real image pairs $[A_y, y]$.
Moreover, we propose the attention-guided adversarial loss for training the attention-guided discriminators.
The min-max game between the attention-guided discriminator $D_{YA}$ and the generator $G$ is performed through the following objective functions:
\begin{equation}
\begin{aligned}
\mathcal{L}_{AGAN}(G, D_{YA}) = &  \mathbb{E}_{y\sim{p_{\rm data}}(y)}\left[ \log D_{YA}([A_y, y])\right]  \\
+ &  \mathbb{E}_{x\sim{p_{\rm data}}(x)}[\log (1 - D_{YA}([A_y, G(x)]))],
\end{aligned}
\label{equ:agan}
\end{equation}
where $D_{YA}$ aims to distinguish between the generated image pairs $[A_y, G(x)]$ and the real image pairs $[A_y,y]$.
We also have another  loss $\mathcal{L}_{AGAN}(F, D_{XA})$ for discriminator $D_{XA}$ and generator $F$, where $D_{XA}$ tries to distinguish the fake image pairs $[A_x, F(y)]$ and the real image pairs $[A_x, x]$.
In this way, the discriminators can focus on the most discriminative content and ignore the unrelated content.

Note that the proposed attention-guided discriminator is only used in scheme I.
In preliminary experiments, we also used the proposed attention-guided discriminator in scheme II, but did not observe  improved performance. 
The reason could be that scheme II has enough ability to learn the most discriminative content between the source and target domains.

\subsection{Optimization Objective}
The optimization objective of the proposed attention-guided generation scheme II can be expressed as:
\begin{equation}
\begin{aligned}
\mathcal{L} = \mathcal{L}_{GAN} + \lambda_{cycle} * \mathcal{L}_{cycle} + \lambda_{id} * \mathcal{L}_{id},
\end{aligned}
\label{eqn:loss}
\end{equation}
where $\mathcal{L}_{GAN}$, $\mathcal{L}_{cycle}$ and $\mathcal{L}_{id}$ are GAN, cycle-consistency, and identity preserving loss~\cite{zhu2017toward}, respectively.
$\lambda_{cycle}$ and $\lambda_{id}$ are parameters controlling the relative relation of each term.
The optimization objective of the proposed attention-guided generation scheme I can be expressed:
\begin{equation}
\begin{aligned}
\mathcal{L} = &  \lambda_{cycle} * \mathcal{L}_{cycle}+ \lambda_{pixel} * \mathcal{L}_{pixel} \\
+  & \lambda_{gan}* (\mathcal{L}_{GAN} + \mathcal{L}_{AGAN}) + \lambda_{tv} * \mathcal{L}_{tv},  
\end{aligned}
\label{eqn:curriculum}
\end{equation}
where $\mathcal{L}_{GAN}$, $\mathcal{L}_{AGAN}$, $\mathcal{L}_{cycle}$, $\mathcal{L}_{tv}$ and $\mathcal{L}_{pixel}$ are GAN, attention-guided GAN, cycle-consistency, attention, and pixel loss, respectively.
$\lambda_{gan}$, $\lambda_{cycle}$, $\lambda_{pixel}$ and $\lambda_{tv}$ are parameters controlling the relative relation of each term.
In the following, we will introduce the attention loss and pixel loss. Note that both losses are only used in the scheme I since the generator needs stronger constraints than those in scheme~II.

When training AttentionGAN, we do not have ground-truth annotation for the attention masks.
They are learned from the resulting gradients of both attention-guided generators and discriminators and the rest of the losses. 
However, the attention masks can easily saturate to 1 causing the attention-guided generators to have no effect. 
To prevent this situation, we perform a Total Variation regularization over attention masks $A_x$ and $A_y$. 
The attention loss of $A_x$ thus can be defined as:
\begin{equation}
	\begin{aligned}
		\mathcal{L}_{tv} =  \sum_{w,h=1}^{W,H} & \left| A_x(w+1, h, c) - A_x(w, h, c)  \right|   \\ 
		+ &\left| A_x(w, h+1, c) - A_x(w, h, c) \right|,
	\end{aligned}
	\label{eqn:tv}
\end{equation}
where $W$ and $H$ are the width and height of $A_x$.

Moreover, to reduce the changes and constrain the generator in scheme I, we adopt the pixel loss between the input images and the generated images.
This loss can be regarded as another form of the identity preserving loss.
We express this loss as:
\begin{equation}
	\begin{aligned}
		\mathcal{L}_{pixel}(G, F) = & \mathbb{E}_{x\sim{p_{\rm data}}(x)}[\Arrowvert G(x)-x\Arrowvert_1]   \\
		+ & \mathbb{E}_{y\sim{p_{\rm data}}(y)}[\Arrowvert F(y)-y\Arrowvert_1].
	\end{aligned}
	\label{equ:pixelloss}
\end{equation}
We adopt $L1$ distance as loss measurement in pixel loss.

\subsection{Implementation Details}
\noindent \textbf{Network Architecture.} For a fair comparison, we use the generator architecture from CycleGAN~\cite{zhu2017unpaired}.
We have slightly modified it for our task. 
Scheme I takes a three-channel RGB image as input and outputs a one-channel attention mask and a three-channel content mask. 
Scheme II takes an three-channel RGB image as input and outputs $n$ attention masks and $n{-}1$ content masks, thus we fuse all of these masks and the input image to produce the final results. 
Specifically, the attention generator in Fig.~\ref{fig:v2} consists of $u128$,
$u64$, and $c7s1{-}10$, where $uk$ denotes a $3 {\times} 3$ fractional-strided-Convolution-InstanceNorm-ReLU layer with $k$ filters and stride $1/2$;  $c7s1{-}k$ denote a $7 {\times}7$ Convolution-InstanceNorm-ReLU layer with $k$ filters and stride 1.
The output of the attention generator is represented by ten attention masks; we choose the first one as the background attention mask, and the rest as the foreground attention masks.
We set $n{=}10$ in our experiments.
For the vanilla discriminator, we employ the discriminator architecture from~\cite{zhu2017unpaired}. 
We employ the same architecture as the proposed attention-guided discriminator except the attention-guided discriminator takes an attention mask and an image as input, while the vanilla discriminator only takes an image as input.

\noindent \textbf{Training Strategy.} 
Our innovation lies in the structure of the proposed model rather than the training approach.
Thus, we follow the standard optimization method from~\cite{goodfellow2014generative,zhu2017unpaired} to optimize the proposed AttentionGAN for fair comparisons, i.e., we alternate between one gradient descent step on the generators, then one step on discriminators.
To slow down the rate of discriminators relative to generators, we follow CycleGAN~\cite{zhu2017unpaired} and divide the objective by 2 while optimizing the discriminators.
We also use a least square loss to stabilize our model during the training procedure, which is more stable than the negative log likelihood objective.
Moreover, we use a history of the generated images to update discriminators similar to CycleGAN.

%% file: 4experiments.tex
\section{Experiments}
\label{sec:experiment}


\subsection{Experimental Setups}
\noindent \textbf{Datasets.}
We follow previous works \cite{choi2017stargan,tang2019attention,kim2019ugatit,zhu2017unpaired} and employ eight popular datasets to evaluate the proposed AttentionGAN, including four face image datasets (i.e., CelebA \cite{liu2015celeba}, RaFD~\cite{langner2010presentation}, AR Face~\cite{martinez1998ar},~and Selfie2Anime~\cite{kim2019ugatit}) and four natural image datasets (i.e., Horse2Zebra~\cite{zhu2017unpaired}, Apple2Orange~\cite{zhu2017unpaired}, Maps~\cite{zhu2017unpaired}, and Style Transfer~\cite{zhu2017unpaired}). More details about these datasets can be found in the corresponding papers.
Note that these eight datasets are widespread for evaluating unsupervised image translation models since they contain diverse scenarios, including natural images, face images, artistic paintings, and aerial photographs. Therefore, it is challenging to built a general and versatile model to handle all of these tasks on various datasets.

\noindent \textbf{Parameter Settings.}
For all datasets, images are rescaled to $256 {\times} 256$. 
We do left-right flip and random crop for data augmentation.
We set the number of image buffer to 50 similar in~\cite{zhu2017unpaired}.
We use the Adam optimizer~\cite{kingma2014adam} with the momentum terms $\beta_1{=}0.5$ and $\beta_2{=}0.999$.
For fair comparisons, we follow \cite{tang2019attention} and set $\lambda_{cycle}{=}10$, $\lambda_{gan}{=}0.5$, $\lambda_{pixel}{=}1$ and $\lambda_{tv}{=}1e{-}6$ in Eq.~\eqref{eqn:curriculum}.
Moreover, we follow \cite{zhu2017unpaired} and set $\lambda_{cycle}{=}10$, $\lambda_{id}{=}0.5$ in Eq.~\eqref{eqn:loss}.
Previous works \cite{tang2019attention,zhu2017unpaired} have verified the effectiveness of these parameters in the unsupervised image-to-image translation task.

\noindent \textbf{Competing Models.}
We consider several state-of-the-art image translation models as our baselines.
1) Unpaired image translation methods: CycleGAN~\cite{zhu2017unpaired}, DualGAN~\cite{yi2017dualgan}, DIAT~\cite{li2016deep}, DiscoGAN~\cite{kim2017learning}, DistanceGAN~\cite{benaim2017one}, Dist.+Cycle~\cite{benaim2017one}, Self Dist.~\cite{benaim2017one}, ComboGAN~\cite{anoosheh2017combogan},  UNIT~\cite{liu2017unsupervised}, MUNIT~\cite{huang2018multimodal}, DRIT~\cite{lee2018diverse}, GANimorph~\cite{gokaslan2018improving}, CoGAN~\cite{liu2016coupled}, SimGAN~\cite{shrivastava2017learning}, Feature loss+GAN~\cite{shrivastava2017learning} (a variant of SimGAN).
2) Paired image translation methods: BicycleGAN~\cite{zhu2017toward}, Pix2pix~\cite{isola2016image}, Encoder-Decoder~\cite{isola2016image}.
3) Class label, object mask, or attention-guided image translation methods: IcGAN~\cite{perarnau2016invertible}, StarGAN~\cite{choi2017stargan}, ContrastGAN~\cite{liang2017generative}, GANimation~\cite{pumarola2018ganimation}, RA~\cite{wang2017residual}, UAIT~\cite{mejjati2018unsupervised}, U-GAT-IT~\cite{kim2019ugatit}, SAT~\cite{yang2019show}, TransGaGa \cite{wu2019transgaga},  DA-GAN \cite{ma2018gan}.
4) Unconditional GANs methods: BiGAN/ALI~\cite{donahue2016adversarial,dumoulin2016adversarially}.
Note that the fully supervised Pix2pix, Encoder-Decoder (Enc.-Decoder) and BicycleGAN are trained on paired data.
Since BicycleGAN can generate several different outputs with one single input image, we randomly select one output from them for fair comparisons. 
To re-implement ContrastGAN, we use OpenFace~\cite{baltruvsaitis2016openface} to obtain the face masks as extra input data.

\noindent \textbf{Evaluation Metrics.}
We follow CycleGAN~\cite{zhu2017unpaired} and adopt Amazon Mechanical Turk (AMT) perceptual studies to evaluate the generated images.
Moreover, to seek a quantitative measure that does not require human participation, Peak Signal-to-Noise Ratio (PSNR), Kernel Inception Distance (KID)~\cite{binkowski2018demystifying} and Fr\'{e}chet Inception Distance (FID)~\cite{heusel2017gans} are employed according to different translation tasks.
Specifically, PSNR measures the pixel-level similarity between the generated image and the real one.
Both KID and FID evaluate the generated images from a high-level feature space.
These three metrics are the most mainstream methods to measure the images generated by unsupervised image translation models.

\subsection{Experimental Results}

\subsubsection{Ablation Study}
We first conduct extensive ablation studies to evaluate the effectiveness of the proposed methods.

\begin{figure}[!t] \footnotesize
	\centering
	\includegraphics[width=0.85\linewidth]{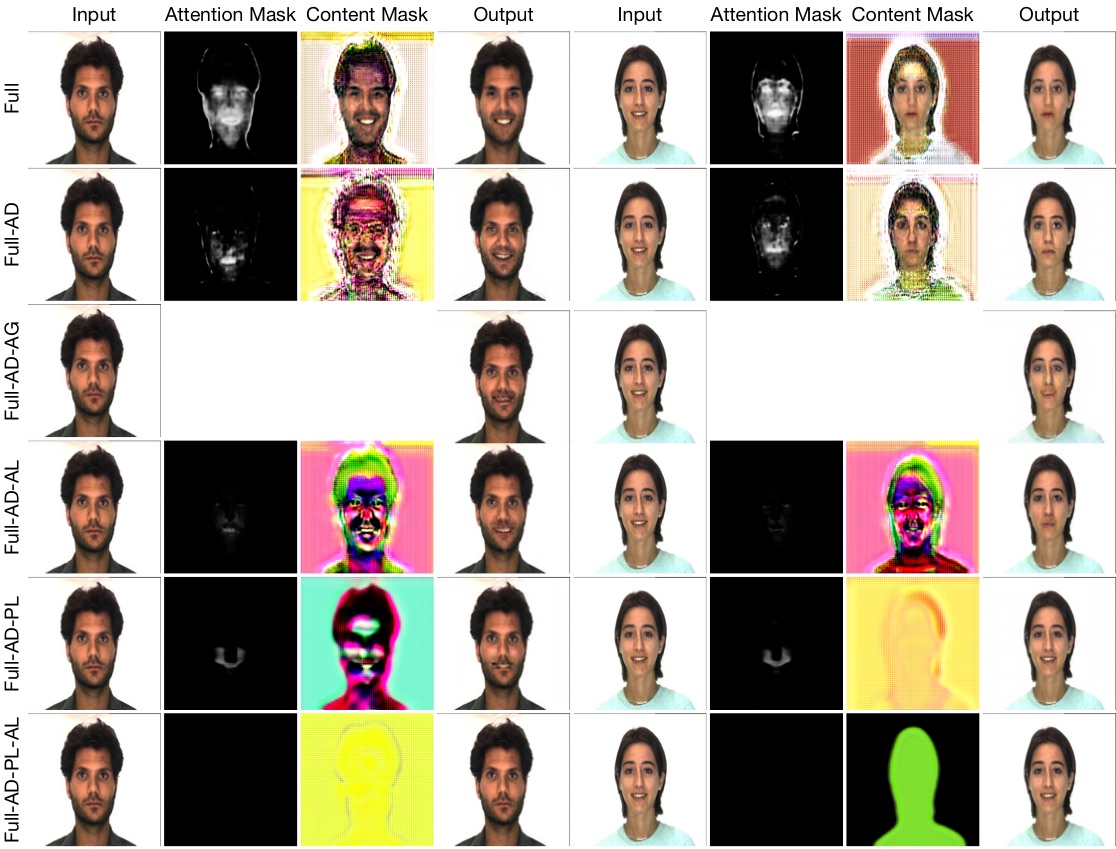}
	\caption{Ablation study of AttentionGAN on AR Face.}
	\label{fig:ar_loss} 
	\vspace{-0.2cm}
\end{figure}

\noindent \textbf{Analysis of Model Component.}
To evaluate the components of our AttentionGAN, we first conduct extensive ablation studies.
We gradually remove components of the proposed AttentionGAN, i.e., Attention-guided Discriminator (AD), Attention-guided Generator (AG), Attention Loss (AL) and Pixel Loss (PL).
The results of AMT and PSNR on AR Face are shown in Table \ref{tab:abl}.
We find that removing one of them substantially degrades the results, which means all of them are critical to our results.
We conduct more experiments on the more challenging Horse2Zebra dataset to validate the effectiveness of each component of the proposed AttentionGAN. The results are shown in Table  \ref{tab:abl}, we can see that the proposed method achieves similar results both on Horse2Zebra and on AR Face datasets.
We also provide qualitative results of AR Face in Fig.~\ref{fig:ar_loss}.
Note that without AG we cannot generate both attention and content masks.

\begin{table}[!t] \small
	\centering
	\caption{Ablation study of AttentionGAN on AR Face and Horse2Zebra.} 
		\begin{tabular}{l|cc|c}
			\multirow{2}{*}{Setup}  & \multicolumn{2}{c|}{AR Face}  & Horse2Zebra \\ \cline{2-4}
			&  AMT $\uparrow$  &   PSNR $\uparrow$                      & FID $\downarrow$                  \\ \midrule
			Full                            & \textbf{12.8} & \textbf{14.9187}  & \textbf{68.55}\\
			Full - AD                   & 10.2              & 14.6352               & 73.63\\
			Full - AD - AG           & 3.2                & 14.4646               & 90.78\\
			Full - AD - PL           &  8.9               &    14.5128              & 78.55\\  
			Full - AD - AL           &  6.3               &   14.6129              &  82.35\\  
			Full - AD - PL - AL    &  5.2               &  14.3287               & 86.14 \\   
	\end{tabular}
	\label{tab:abl}
	\vspace{-0.4cm}
\end{table}

\begin{figure}[!t] \small
	\centering
	\includegraphics[width=0.845\linewidth]{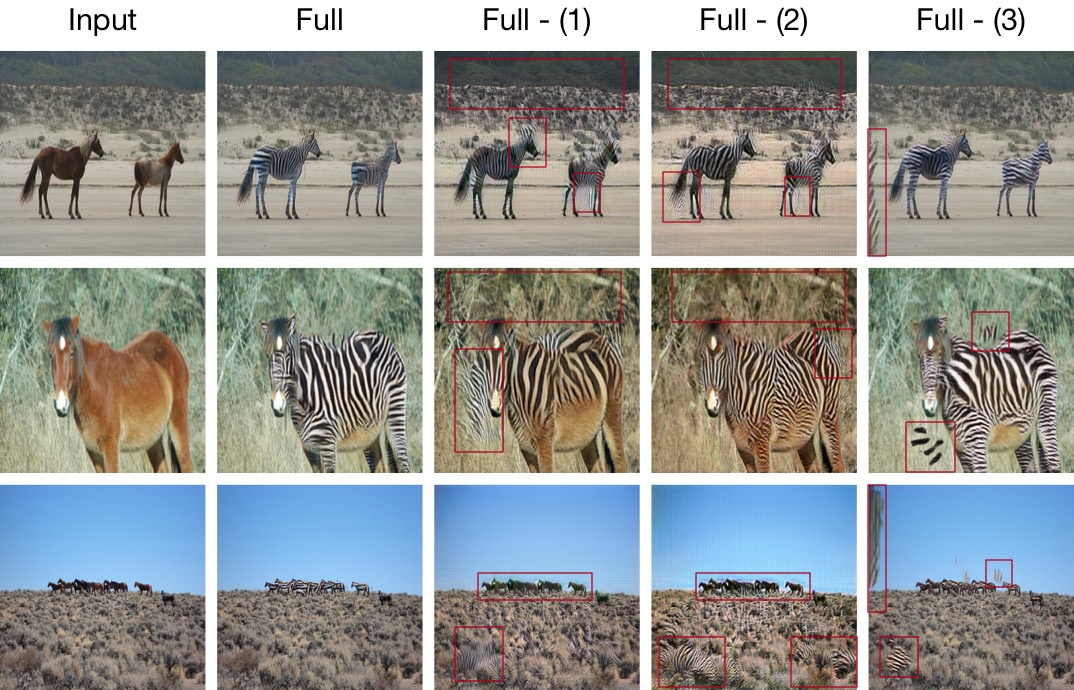}
	\caption{Ablation study of AttentionGAN on Horse2Zebra.}
	\label{fig:abl2} 
	\vspace{-0.4cm}
\end{figure}

We conduct more ablation studies to verify how much of the improvement is due to the three designs: (1) employing two separate sub-networks to generate content mask and attention mask. (2) producing both foreground mask and background mask. (3) learning a set of intermediate content masks. 
Results are shown in Table \ref{tab:abl2} and Fig. \ref{fig:abl2}, where `Full' denotes our full model, `Full-(1)' denotes that we use a shared network to generate both content mask and attention mask, `Full-(2)' denotes that we only produce foreground masks, `Full-(3)' denotes that we learn only one content mask.
As can be seen in Table \ref{tab:abl2} that removing one of these three designs significantly decreases the generation performance, validating all of them are critical to learn a better unsupervised image-to-image translation mapping.
Moreover, `Full' generates much better results than others (see Fig. \ref{fig:abl2}). Specifically, `Full-(3)' generates a lot of visual artifacts (shown in the red boxes), but `Full' does not. `Full-(1)' and `Full-(2)' change the background while translating, but `Full' does not. In addition, both `Full-(1)' and `Full-(2)' also produce some artifacts, which make the generated zebras look blurry (see the red boxes near the zebras of both `Full-(1)' and `Full-(2)').

Lastly, we observe that ``Full-(1)'' in Table \ref{tab:abl2} performs much worse ``Full-AD-AG'' in Table \ref{tab:abl}. ``Full-(1)'' denotes that we use a shared network to generate both content mask and attention mask.
Since there is a big difference between attention mask (i.e., grayscale image) and content mask (i.e., RGB image), if we use one network to learn both, the two of them will pollute each other, causing both to learn poorly.
In this case, the result of ``Full-(1)'' will be worse than that of ``Full-AD-AG''.
This means that the design ``(1)'' is a prerequisite for both designs ``(2)'' and ``(3)'', and the design ``(1)'' is also the most important design among the three. In the end, the performance will be the best when only the three are used together (see Table \ref{tab:abl2}).

\begin{figure}[!t] \footnotesize
	\centering
	\includegraphics[width=0.84\linewidth]{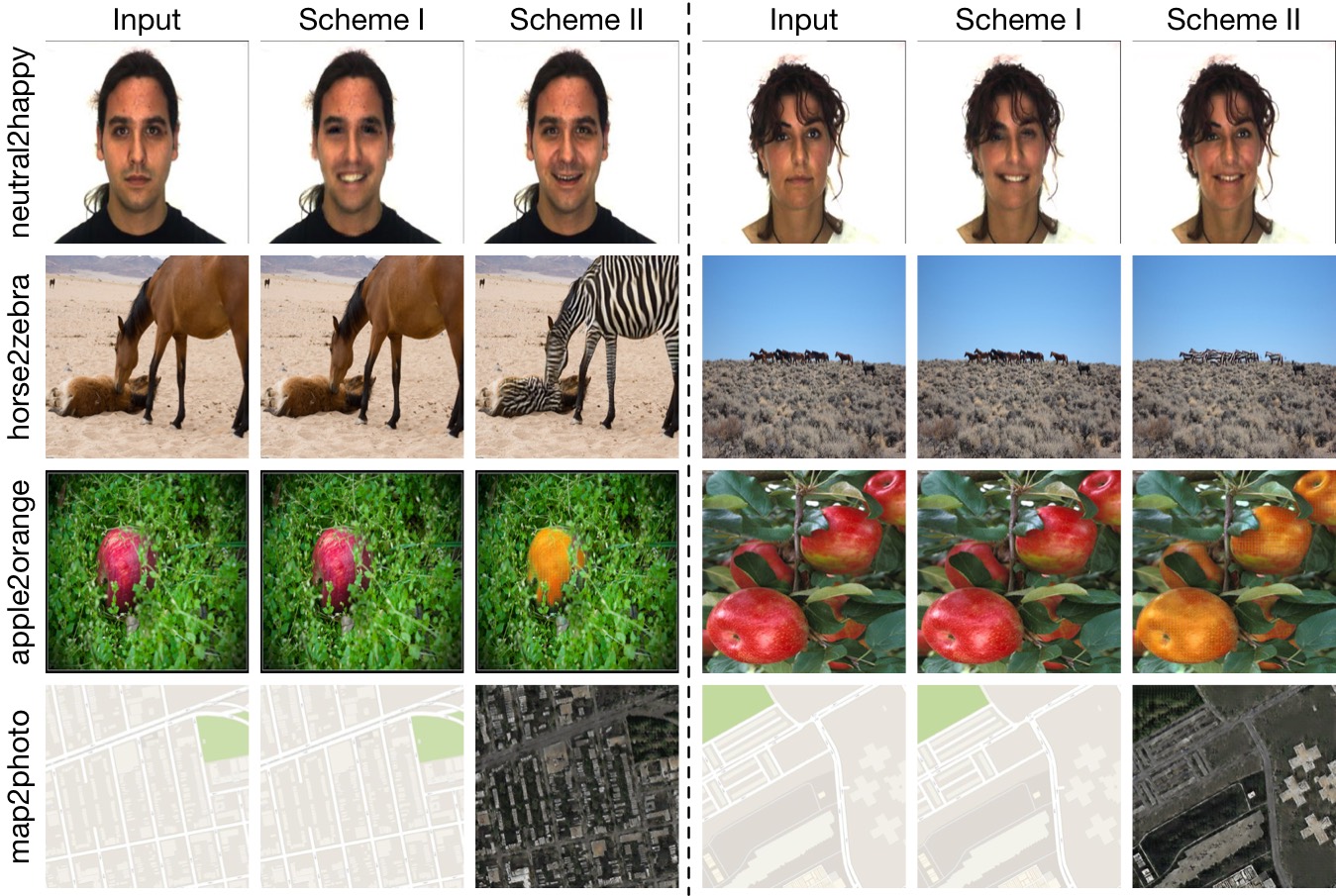}
	\caption{Comparison results of the proposed attention-guided generation scheme I and II.}
	\label{fig:scheme12} 
\end{figure}

\begin{table}[!t] \small
	\centering
	\caption{Ablation study of AttentionGAN on Horse2Zebra.}
		\begin{tabular}{lcccc} 		    
			Setup                      & Full       & Full-(1)           & Full-(2)                   & Full-(3) \\ \midrule
			FID $\downarrow$   & \textbf{68.55}   & 101.80            & 77.29                      & 79.33 \\
	\end{tabular}
	\label{tab:abl2}
	\vspace{-0.4cm}
\end{table}

\noindent \textbf{Attention-Guided Generation Scheme I vs. II}
Moreover, we present the comparison results of the proposed attention-guided generation schemes I and II.
Scheme I is used in our conference paper~\cite{tang2019attention}. 
Scheme II is a refined version proposed in this paper. 
Comparison results are shown in Fig.~\ref{fig:scheme12}.
We observe that scheme I generates good results on facial expression transfer, however, it generates  identical images with the inputs on other tasks, e.g., horse to zebra translation, apple to orange translation, and map to aerial photo translation.
However, the proposed attention-guided generation scheme II can handle all of these tasks, which validates the effectiveness of the proposed scheme II.

\begin{figure}[!t] \footnotesize
	\centering
	\includegraphics[width=0.6\linewidth]{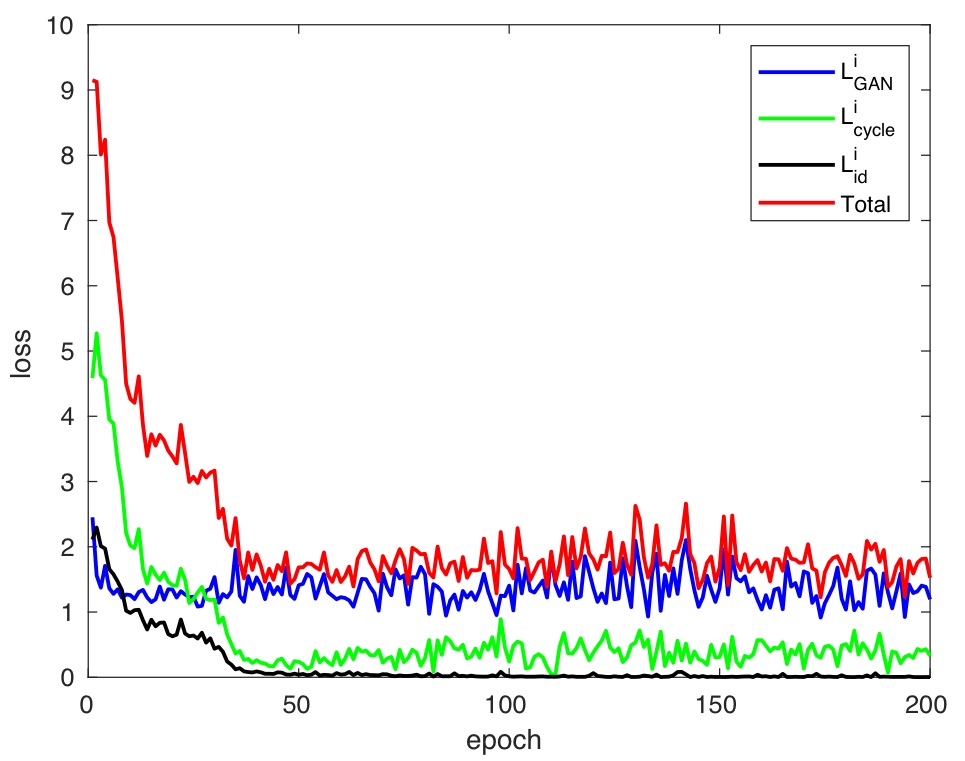}
	\caption{Convergence loss in Eq.~\eqref{eqn:loss} on Horse2Zebra.}
	\label{fig:loss}
	\vspace{-0.2cm}
\end{figure}

\noindent \textbf{Analysis of Hyper-Parameters.}
We investigate the influence of $\lambda_{cycle}$ and $\lambda_{id}$ in Eq.~\eqref{eqn:loss} on the performance of our method. 
As can be seen in Table \ref{tab:para_cyc} and \ref{tab:para_id}, when $\lambda_{cycle}{=}10$ and $\lambda_{id}{=}0.5$, the proposed method achieves the best FID score on Horse2Zebra, validating the effectiveness of our setting of hyper-parameters.
Moreover, to show that the proposed framework can stabilize GAN training, we illustrate the
convergence loss of the proposed method in Eq.~\eqref{eqn:loss} on Horse2Zebra (see Fig.~\ref{fig:loss}).
We observe that the proposed method ensures a very fast yet stable convergence (around the 50th epoch).

\subsubsection{Experiments on Face Images}
We conduct extensive experiments on four datasets to validate AttentionGAN.

\begin{table}[!t] \small
	\centering
	\caption{The influence of $\lambda_{cycle}$ on Horse2Zebra.}
	\begin{tabular}{lccccc} 		    
		$\lambda_{cycle}$                    & 0.1       & 1           & 10                   & 100        & 1000 \\ \midrule
		FID $\downarrow$                    & 137.74 &  79.10   & \textbf{68.55} & 244.27   & 244.22 \\
	\end{tabular}
	\label{tab:para_cyc}
	\vspace{-0.2cm}
\end{table}

\begin{table}[!t] \small
	\centering
	\caption{The influence of $\lambda_{id}$ on Horse2Zebra.}
	\begin{tabular}{lccccc} 		    
		$\lambda_{id}$                          & 0.01      & 0.1     & 0.5      & 1          & 10 \\ \midrule
		FID $\downarrow$                     & 106.76  & 83.67 & \textbf{68.55}  & 117.97  & 244.48 \\
	\end{tabular}
	\label{tab:para_id}
	\vspace{-0.4cm}
\end{table}

\begin{figure}[!t] \footnotesize
	\centering
	\includegraphics[width=1\linewidth]{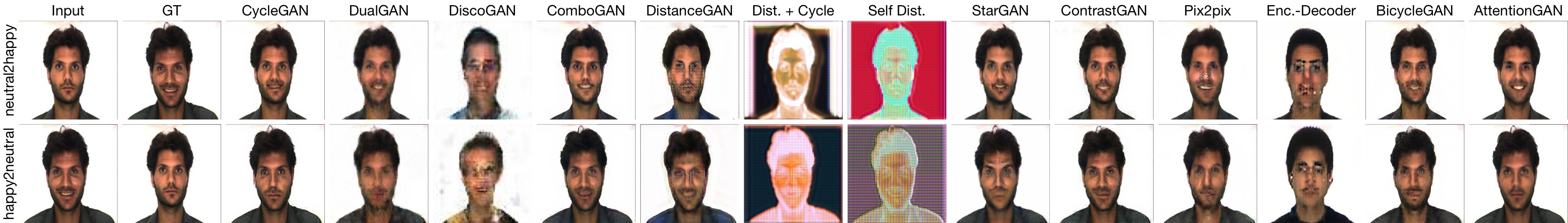}
	\caption{Results of facial expression transfer on AR Face.}
	\label{fig:bu3dfe_ar}
	\vspace{-0.4cm}
\end{figure}

\noindent \textbf{Results on AR Face.}  
The results of neutral $\leftrightarrow$ happy expression translation on AR Face are shown in Fig.~\ref{fig:bu3dfe_ar}.
Clearly, the results of Dist.+Cycle and Self Dist. cannot even generate human faces.
DiscoGAN produces identical results regardless of the input faces suffering from mode collapse.
The results of DualGAN, DistanceGAN, StarGAN, Pix2pix, Encoder-Decoder and BicycleGAN tend to be blurry, while ComboGAN and ContrastGAN produce the same identity but without expression changing.
CycleGAN generates sharper images, but the details of the generated faces are not convincing.
Compared with all the baselines, the results of AttentionGAN are correct, smoother, and with more details.

\begin{figure}[!t] \footnotesize
	\centering
	\includegraphics[width=1\linewidth]{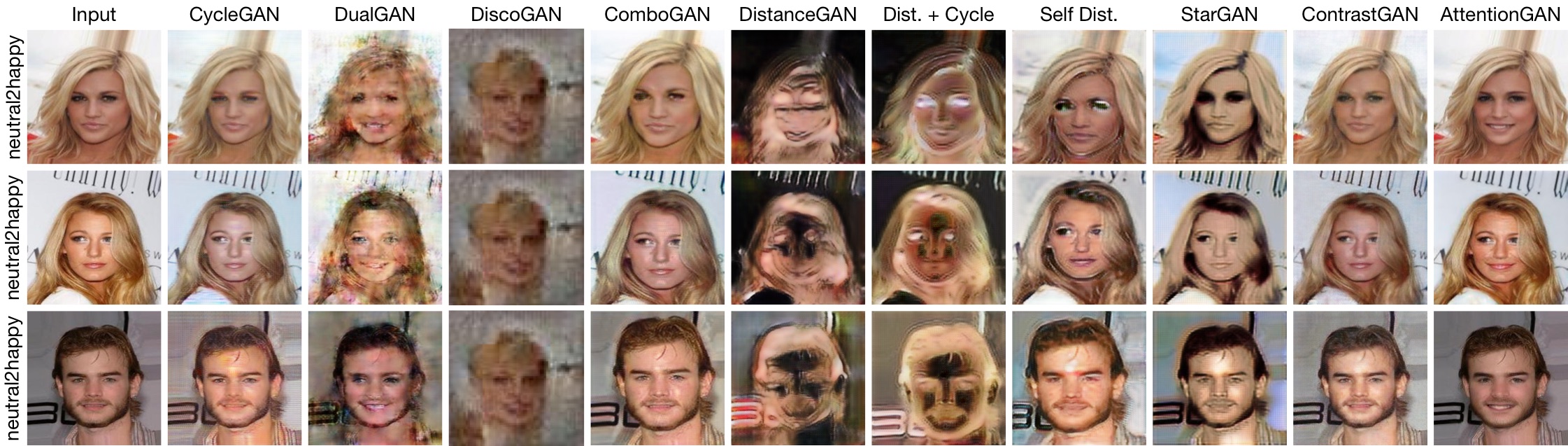}
	\caption{Results of facial expression transfer on CelebA.}
	\label{fig:celeba}
	\vspace{-0.4cm}
\end{figure}

\noindent \textbf{Results on CelebA.}
We conduct both facial expression translation and facial attribute transfer tasks on this dataset.
Facial expression translation task on this dataset is more challenging than AR Face dataset since the background of this dataset is very complicated.
Note that this dataset does not provide paired data, thus we cannot conduct experiments with supervised methods, i.e., Pix2pix, BicycleGAN, and Encoder-Decoder.
The results compared with other baselines are shown in Fig.~\ref{fig:celeba}. 
We observe that only the proposed AttentionGAN produces photo-realistic faces with correct expressions.
The reason could be that methods without attention cannot learn the most discriminative part and the unwanted part.
All existing methods failed to generate novel expressions, which means they treat the whole image as the unwanted part, while the proposed AttentionGAN can learn novel expressions by distinguishing the discriminative part from the unwanted part.

\begin{figure}[!t] \footnotesize
	\centering
	\includegraphics[width=1\linewidth]{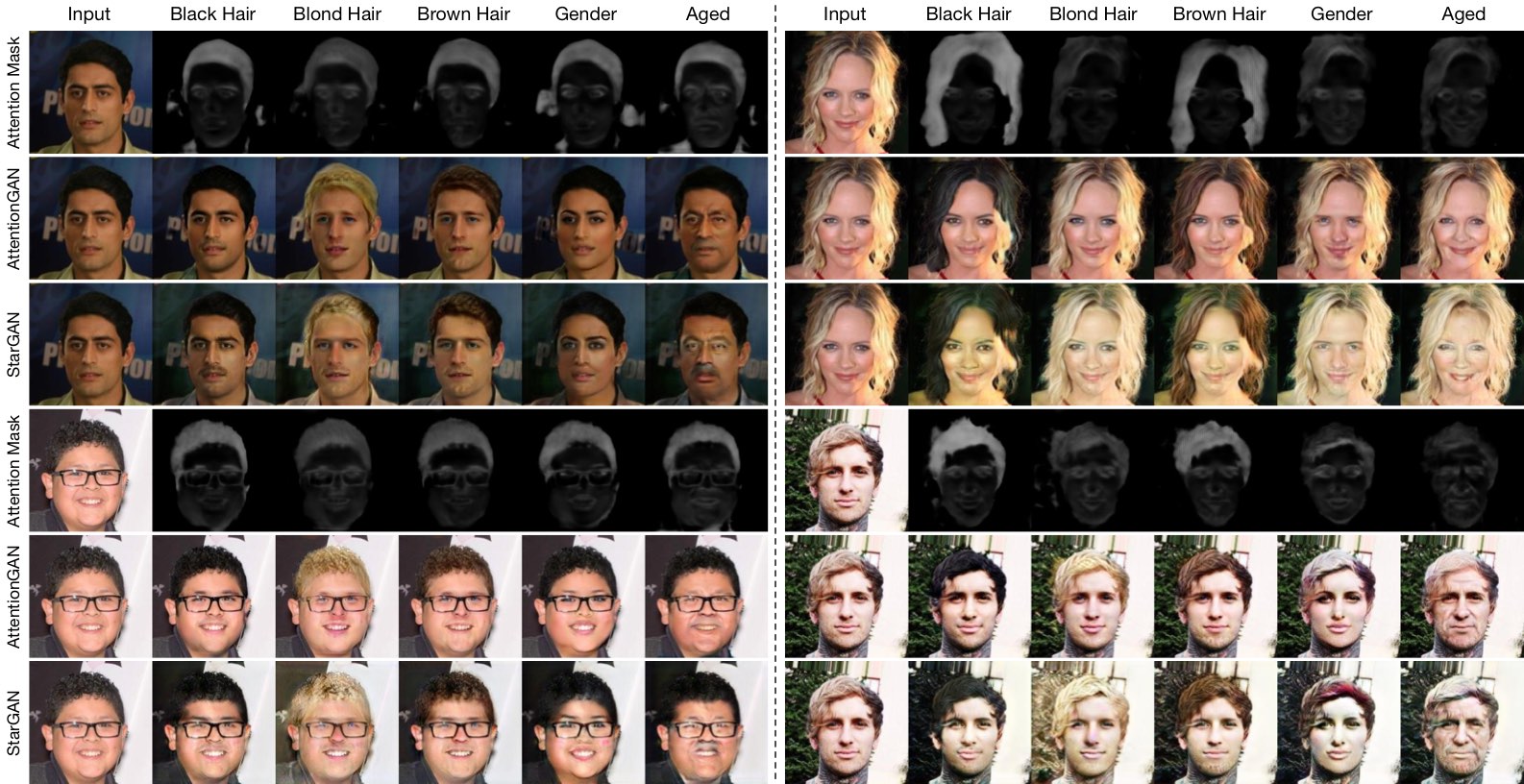}
	\caption{Results of facial attribute transfer on CelebA.}
	\label{fig:celeba_com}
	\vspace{-0.4cm}
\end{figure}

Moreover, our model can be easily extended to solve multi-domain image-to-image translation problems.
To control multiple domains in one single model we employ the domain classification loss proposed in StarGAN. 
Thus, we follow StarGAN and conduct facial attribute transfer on this dataset to evaluate the proposed AttentionGAN.
The results compared with StarGAN are shown in Fig.~\ref{fig:celeba_com}.
We observe that the proposed AttentionGAN achieves visually better results than StarGAN without changing background.

\begin{figure}[!t] \footnotesize
	\centering
	\includegraphics[width=1\linewidth]{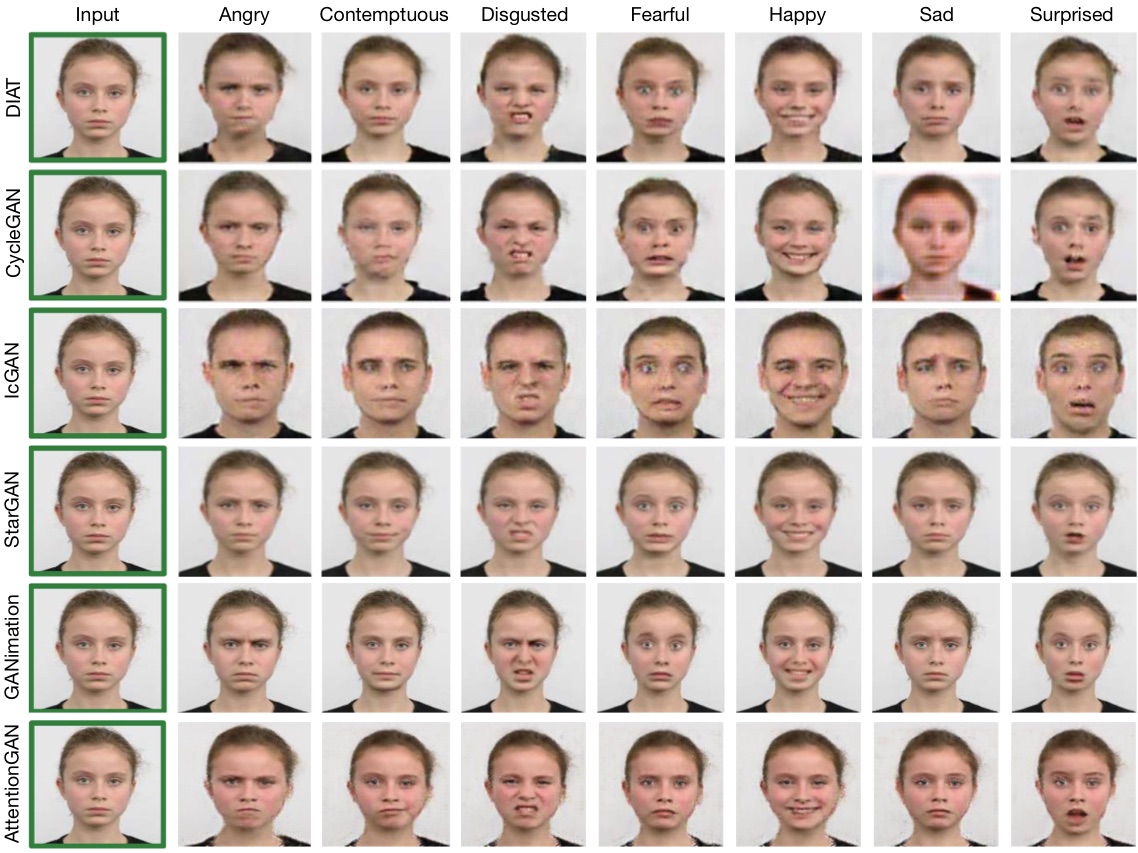}
	\caption{Results of facial expression transfer on RaFD.}
	\label{fig:rafd}
	\vspace{-0.4cm}
\end{figure}

\noindent \textbf{Results on RaFD.} 
We follow StarGAN and conduct diversity facial expression translation on this dataset.
The results compared to the baselines DIAT, CycleGAN, IcGAN, StarGAN, and GANimation are shown in Fig.~\ref{fig:rafd}.
We observe that the proposed AttentionGAN achieves better results than DIAT, CycleGAN, StarGAN, and  IcGAN.
For GANimation, we follow the authors' instructions and use OpenFace~\cite{baltruvsaitis2016openface} to obtain the action units of each face as extra input data.
Note that the proposed method generates competitive results compared to GANimation but GANimation needs action units annotations as extra training data limiting its practical application.
More importantly, GANimation cannot handle other generative tasks such facial attribute transfer as shown in Fig.~\ref{fig:celeba_com}.

\begin{figure}[!t] \footnotesize
	\centering
	\includegraphics[width=0.85\linewidth]{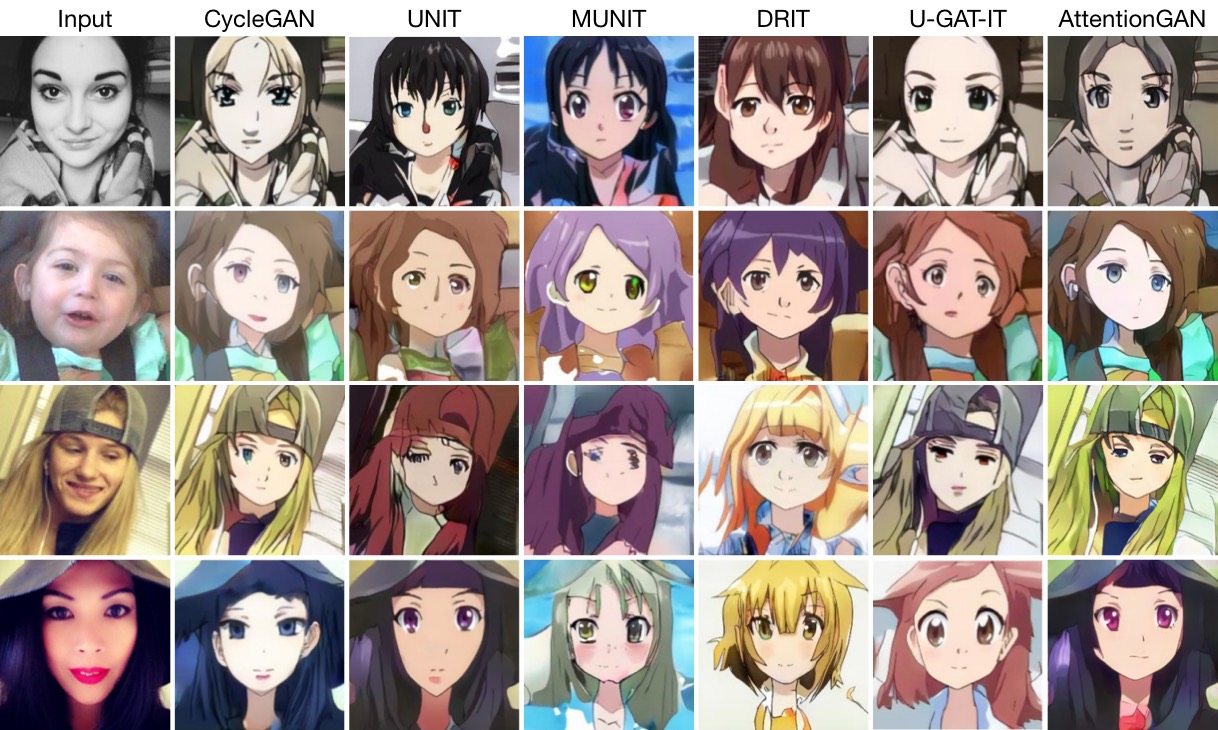}
	\caption{Different methods for mapping selfie to anime.}
	\label{fig:se} 
	\vspace{-0.4cm}
\end{figure}

\noindent \textbf{Results of Selfie2Anime.} 
We follow U-GAT-IT~\cite{kim2019ugatit} and conduct selfie to anime translation on the  Selfie2Anime dataset.
The results compared with state-of-the-art methods are shown in Fig.~\ref{fig:se}.
We observe that the proposed AttentionGAN achieves better results than other baselines.

We conclude that even though the subjects in these four datasets have different races, poses, styles, skin colors, illumination conditions, occlusions, and complex backgrounds, our method consistently generates more sharper images with correct expressions/attributes than existing models. 
We also observe that our AttentionGAN performs better than other baselines when training data are limited (see Fig.~\ref{fig:celeba}), which also shows that our method is very robust.

\noindent \textbf{Quantitative Comparison.}
We also provide quantitative results of these tasks.
As shown in Table~\ref{tab:result}, AttentionGAN achieves the best results compared with competing models including fully supervised methods (e.g., Pix2pix, Encoder-Decoder and BicycleGAN) and mask-conditional methods (e.g., ContrastGAN).
Next, following StarGAN, we perform a user study using Amazon Mechanical Turk (AMT) to assess attribute transfer task on CelebA.
The results comparing the state-of-the-art methods are shown in Table~\ref{tab:celeba_amt}.
We observe that AttentionGAN achieves significantly better results than all the baselines.
Moreover, we follow U-GAT-IT~\cite{kim2019ugatit} and adopt KID to evaluate the generated images on selfie to anime translation.
The results are shown in Table~\ref{tab:selfie2anime}, we observe that our AttentionGAN achieves the best results compared with baselines except U-GAT-IT.
However, U-GAT-IT needs to adopt two auxiliary classifiers to obtain attention makes.
Moreover, U-GAT-IT uses four discriminators (i.e., two global discriminators and two local discriminators), while we only use two discriminators.
Both designs of U-GAT-IT significantly increase the number of network parameters and training time (see Table \ref{tab:computational}).
Moreover, compared with scheme I, scheme II only adds a little amount of parameters (0.9M), but the performance is significantly improved (see Fig.~\ref{fig:scheme12}). This indicates that the performance improvement of scheme II does not come from simply increasing network parameters.

\begin{table}[!t] \small
	\centering
	\caption{Quantitative comparison on the facial expression translation task. For both AMT and PSNR, high is better. }
		\begin{tabular}{r|cc|c} 
			\multirow{2}{*}{Method}  & \multicolumn{2}{c|}{AR Face} & CelebA \\ \cline{2-4}
			 & AMT $\uparrow$ & PSNR $\uparrow$                     & AMT $\uparrow$   \\ \midrule		
			CycleGAN~\cite{zhu2017unpaired}  			  & 10.2  & 14.8142  & 34.6     \\ 
			DualGAN~\cite{yi2017dualgan}       			     & 1.3    & 14.7458  & 3.2       \\ 
			DiscoGAN~\cite{kim2017learning}                & 0.1    & 13.1547    & 1.2        \\ 
			ComboGAN~\cite{anoosheh2017combogan} & 1.5     & 14.7465  & 9.6       \\   
			DistanceGAN~\cite{benaim2017one}             & 0.3     & 11.4983          & 1.9     \\  
			Dist.+Cycle~\cite{benaim2017one}                & 0.1     & 3.8632        & 1.3    \\  
			Self Dist.~\cite{benaim2017one}                   & 0.1     & 3.8674       & 1.2       \\  
			StarGAN~\cite{choi2017stargan}                  & 1.6     & 13.5757        & 14.8    \\ 
			ContrastGAN~\cite{liang2017generative}      & 8.3     & 14.8495       & 25.1     \\ 
			Pix2pix~\cite{isola2016image}                      & 2.6      & 14.6118    & -           \\ 
			Enc.-Decoder~\cite{isola2016image}            & 0.1     & 12.6660       & -        \\  
			BicycleGAN~\cite{zhu2017toward}               & 1.5     & 14.7914             & -         \\ 
			AttentionGAN                                             & \textbf{12.8} & \textbf{14.9187}  & \textbf{38.9}  \\ 		
		\end{tabular}
		\vspace{-0.4cm}
		\label{tab:result}
	\end{table}
	
\begin{table}[!t] \small
\centering
\caption{AMT results of the facial attribute transfer task on CelebA. For this metric, higher is better.}
	\begin{tabular}{rccc} 		    
		Method                                      & Hair Color  & Gender & Aged \\ \midrule
		DIAT~\cite{li2016deep}               & 3.5         & 21.1   & 3.2 \\
		CycleGAN~\cite{zhu2017unpaired}     & 9.8         & 8.2    & 9.4 \\
		IcGAN~\cite{perarnau2016invertible}  & 1.3         & 6.3    & 5.7 \\
		StarGAN~\cite{choi2017stargan}        & 24.8        & 28.8   & 30.8 \\
		AttentionGAN                           & \textbf{60.6}        & \textbf{35.6}   & \textbf{50.9} \\
	\end{tabular}
	\label{tab:celeba_amt}
	\vspace{-0.4cm}
\end{table}
		
\begin{table}[!t] \small
\centering
\caption{KID $\times$ 100 $\pm$ std. $\times$ 100 of the selfie to anime translation task. For this metric, lower is better.
}
	\begin{tabular}{rc} 		    
		Method                                      & Selfie to Anime \\ \midrule
		U-GAT-IT~\cite{kim2019ugatit}          & \textbf{11.61 $\pm$ 0.57} \\
		CycleGAN~\cite{zhu2017unpaired}      & 13.08 $\pm$ 0.49 \\		
		UNIT~\cite{liu2017unsupervised}         & 14.71 $\pm$ 0.59 \\
		MUNIT~\cite{huang2018multimodal}    & 13.85 $\pm$ 0.41  \\
		DRIT~\cite{lee2018diverse}                & 15.08 $\pm$ 0.62 \\			
		AttentionGAN                                   & 12.14 $\pm$ 0.43
	\end{tabular}
	\label{tab:selfie2anime}
	\vspace{-0.4cm}
\end{table}

\begin{figure*}[!t] \footnotesize
	\centering
	\includegraphics[width=0.85\linewidth]{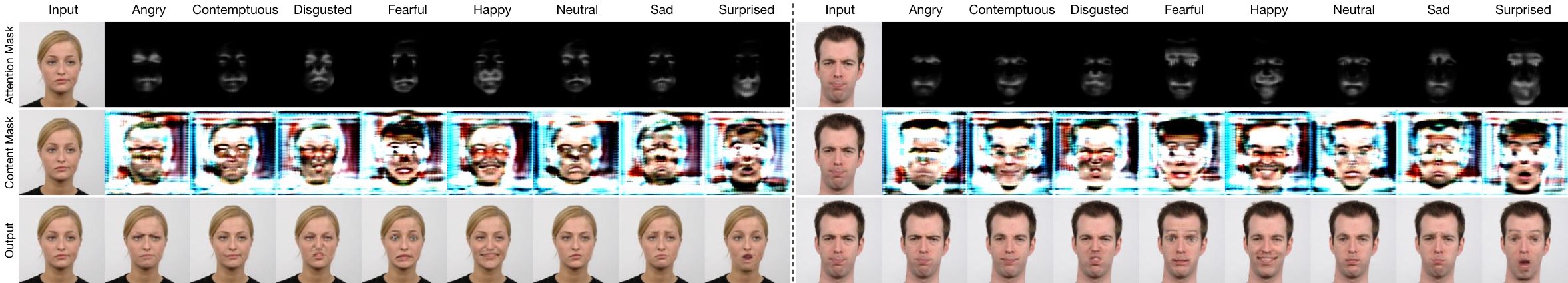}
	\caption{Attention and content masks on RaFD.}
	\label{fig:attention_facial_rafd} 
	\vspace{-0.4cm}
\end{figure*}

\begin{figure*}[!t] \footnotesize
	\centering
	\includegraphics[width=0.85\linewidth]{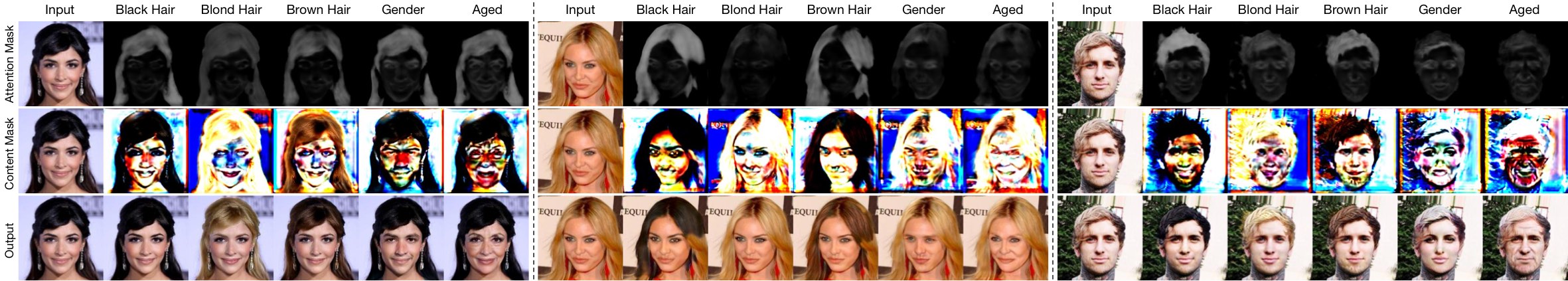}
	\caption{Attention and content masks on CelebA.}
	\label{fig:attention_facial_celeba} 
	\vspace{-0.4cm}
\end{figure*}

\begin{figure*}[!t] \footnotesize
	\centering
	\includegraphics[width=0.85\linewidth]{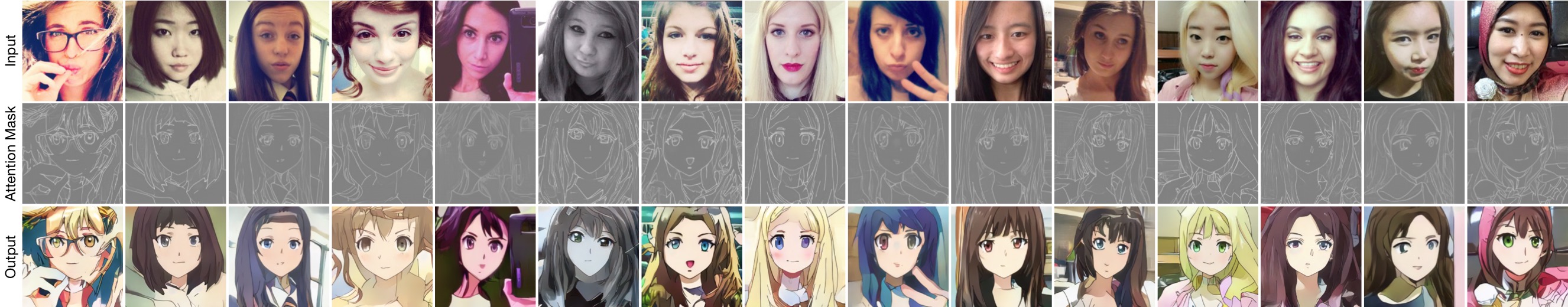}
	\caption{Attention mask on the selfie to anime translation task.}
	\label{fig:attention_selfie} 
	\vspace{-0.4cm}
\end{figure*}

\noindent \textbf{Visualization of Learned Attention and Content Masks.}
Instead of regressing a full image, our generator outputs two masks, a content mask and an attention mask. 
We also visualize both masks on RaFD and CelebA in Fig.~\ref{fig:attention_facial_rafd} and Fig.~\ref{fig:attention_facial_celeba}, respectively. 
In Fig.~\ref{fig:attention_facial_rafd}, we observe that different expressions generate different attention masks and content masks.
The proposed method makes the generator focus only on those discriminative regions of the image that are responsible of synthesizing the novel expression. 
The attention masks mainly focus on the eyes and mouth, which means these parts are important for generating novel expressions. 
The proposed method also keeps the other elements of the image or unwanted parts untouched. 
In Fig.~\ref{fig:attention_facial_rafd}, the unwanted part are hair, cheeks, clothes, and background, which means these parts have no contribution in generating novel expressions.
In Fig.~\ref{fig:attention_facial_celeba}, we observe that different facial attributes also generate different attention masks and content masks, which further validates our initial motivations. More attention masks generated by AttentionGAN on the facial attribute transfer task are shown in Fig.~\ref{fig:celeba_com}.
Note that the proposed AttentionGAN can handle the geometric changes between source and target domains, such as selfie to anime translation. 
Therefore, we show the learned attention masks on selfie to anime translation task to interpret the generation process in Fig.~\ref{fig:attention_selfie}.

\begin{figure}[!t] \footnotesize
	\centering
	\includegraphics[width=0.85\linewidth]{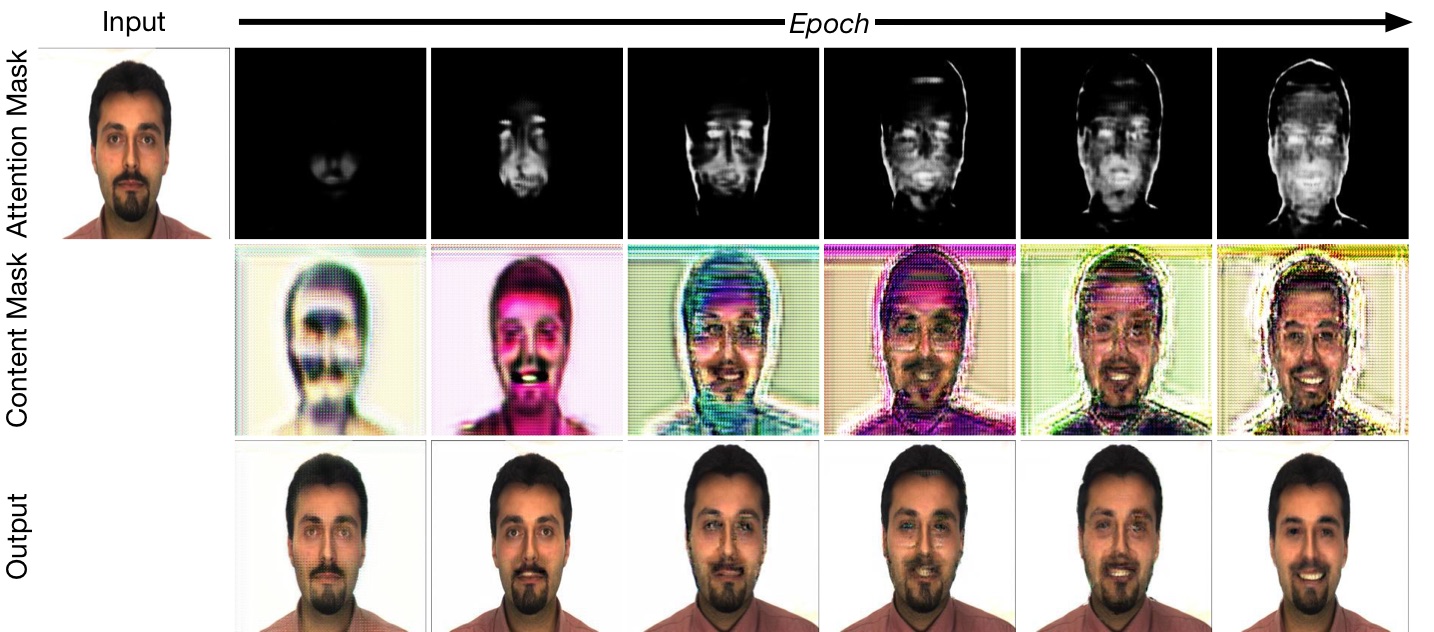}
	\caption{Evolution of attention masks and content masks.}
	\label{fig:ar_model} 
	\vspace{-0.4cm}
\end{figure}

We also present the generation of both attention and content masks on AR Face dataset epoch-by-epoch in Fig.~\ref{fig:ar_model}.
We see that with the number of training epoches increases, the attention mask and the result become better, and the attention masks correlate well with image quality, which demonstrates the proposed AttentionGAN is effective.

\noindent \textbf{Comparison of Model Parameters.}
The number of models for $m$ image domains and the number of model parameters on RaFD are shown in Table~\ref{tab:computational}.
Note that our generation performance is much better than these baselines and the number of parameters is also comparable with ContrastGAN, while ContrastGAN requires object masks as extra data.

\subsubsection{Experiments on Natural Images}
We also conduct experiments on four datasets to evaluate AttentionGAN.

\begin{figure}[!t] \footnotesize
	\centering
	\includegraphics[width=1\linewidth]{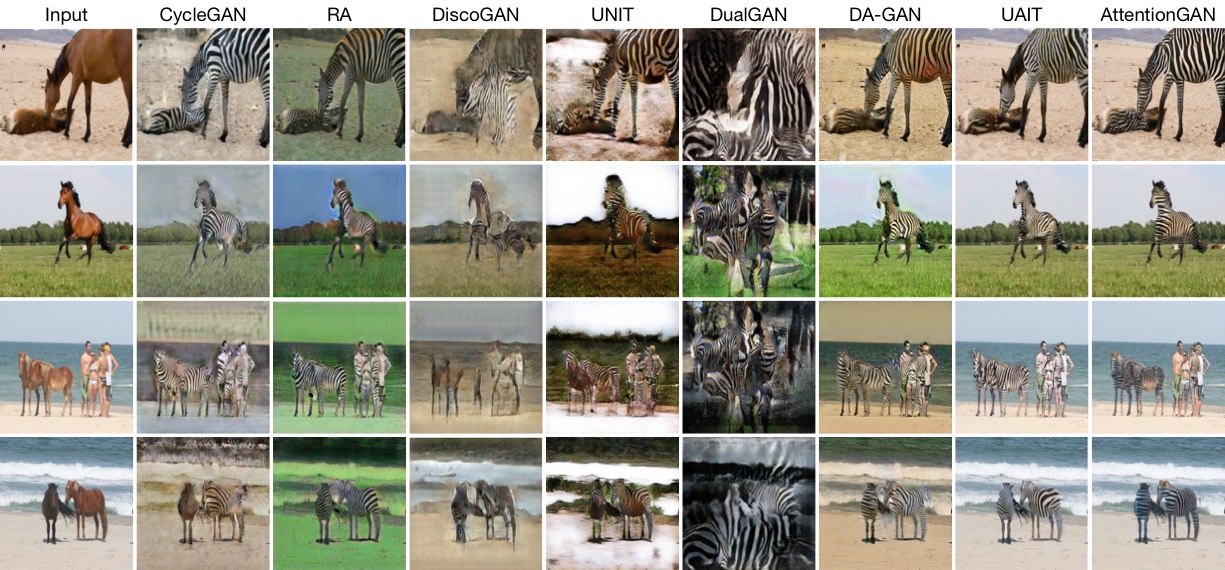}
	\caption{Different methods for mapping horse to zebra.}
	\label{fig:h2z1}
	\vspace{-0.4cm}
\end{figure}

\noindent \textbf{Results of Horse2Zebra.}
The results of horse to zebra translation compared with CycleGAN, RA, DiscoGAN, UNIT, DualGAN, DA-GAN, and UAIT are shown in Fig.~\ref{fig:h2z1}.
We observe that DiscoGAN, UNIT, DualGAN generate blurred results.
CycleGAN, RA, and DA-GAN can generate the corresponding zebras, however the background of images produced by these models has also been changed.
Both UAIT and the proposed AttentionGAN generate the corresponding zebras  without changing the background.
By carefully examining the translated images from both UAIT and the proposed AttentionGAN, we observe that AttentionGAN achieves slightly better results than UAIT as shown in the first and the third rows of Fig.~\ref{fig:h2z1}.
Our method produces better stripes on the body of the lying horse than UAIT as shown in the first row.
In the third row, the proposed method generates fewer stripes on the body of the people than UAIT.

\begin{table}[!t] \small
	\centering
	\caption{Overall model capacity on RaFD ($m{=}8$).}
		\begin{tabular}{rcc} 		    
			Method                & \# Models   & \# Parameters \\ \midrule
			Pix2pix~\cite{isola2016image}                      & m(m-1)  & 57.2M$\times$56 \\
			Encoder-Decoder~\cite{isola2016image} & m(m-1) &  41.9M$\times$56 \\
			BicycleGAN~\cite{zhu2017toward}  & m(m-1)  & 64.3M$\times$56 \\ \hline   
			CycleGAN~\cite{zhu2017unpaired}  & m(m-1)/2        & 52.6M$\times$28 \\
			DualGAN~\cite{yi2017dualgan}  & m(m-1)/2         & 178.7M$\times$28 \\
			DiscoGAN~\cite{kim2017learning}  & m(m-1)/2        & 16.6M$\times$28 \\
			DistanceGAN~\cite{benaim2017one}  & m(m-1)/2   & 52.6M$\times$28 \\
			Dist.+Cycle~\cite{benaim2017one}  & m(m-1)/2   & 52.6M$\times$28 \\
			Self Dist.~\cite{benaim2017one}  & m(m-1)/2   & 52.6M$\times$28 \\ 
		    U-GAT-IT~\cite{kim2019ugatit} & m(m-1)/2 & 134.0M$\times$28\\ \hline
			ComboGAN~\cite{anoosheh2017combogan}  & m                 & 14.4M$\times$8 \\ \hline
			StarGAN~\cite{choi2017stargan} & 1                 & 53.2M$\times$1 \\
			ContrastGAN~\cite{liang2017generative}  & 1          & 52.6M$\times$1 \\ 
			AttentionGAN (Scheme I) \cite{tang2019attention}   & 1           & 52.6M$\times$1 \\ 
			AttentionGAN (Scheme II)     & 1           & 53.5M$\times$1  \\ 
		\end{tabular}
		\label{tab:computational}
		\vspace{-0.4cm}
\end{table}

\begin{figure}[!t] \footnotesize
	\centering
	\includegraphics[width=1\linewidth]{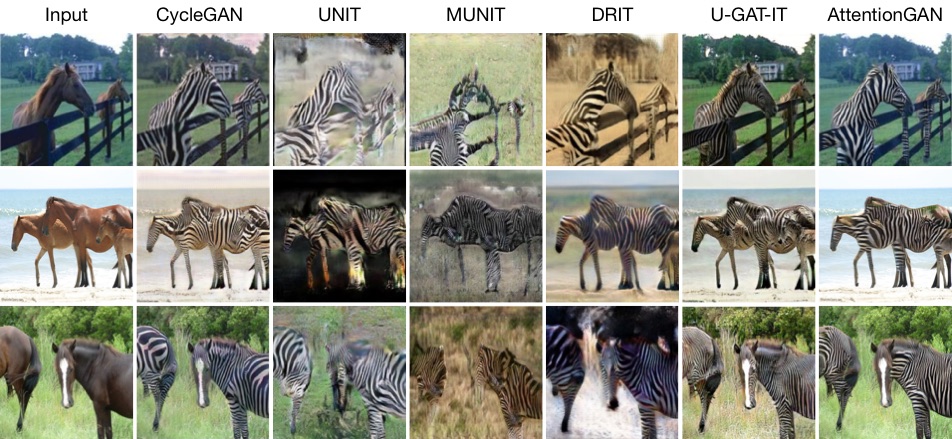}
	\caption{Different methods for mapping horse to zebra.}
	\label{fig:h2z2}
	\vspace{-0.4cm}
\end{figure}

Moreover, we compare the proposed method with CycleGAN, UNIT, MUNIT, DRIT, and U-GAT-IT in Fig.~\ref{fig:h2z2}.
We can see that UNIT, MUNIT and DRIT generate blurred images with many visual artifacts.
CycleGAN can produces the corresponding zebras, however the background of the images has also been changed. 
U-GAT-IT and AttentionGAN can produce better results than other approaches.
However, if we look closely at the results generated by both methods, we observe that U-GAT-IT slightly changes the background, while the proposed  AttentionGAN perfectly keeps the background unchanged.
For instance, as can be seen from the results of the first line, U-GAT-IT produces a darker background than the background of the input image in Fig.~\ref{fig:h2z2}.
While the background color of the generated images by U-GAT-IT is lighter than the input images as shown in the second and third rows in Fig.~\ref{fig:h2z2}.
Lastly, our method behaves the same as CycleGAN and U-GAT-IT, which greatly changes the wooden fence in the first row.
The reason is that part of the horse is occluded by the fence. Our model tries to learn and translate a complete horse, thus it will change the fence.
A feasible solution is to introduce an additional object mask \cite{liang2017generative} or instance mask \cite{mo2018instagan} to alleviate this phenomenon, which will be further investigated in our future work.

\begin{figure}[!t] \footnotesize
	\centering
	\includegraphics[width=1\linewidth]{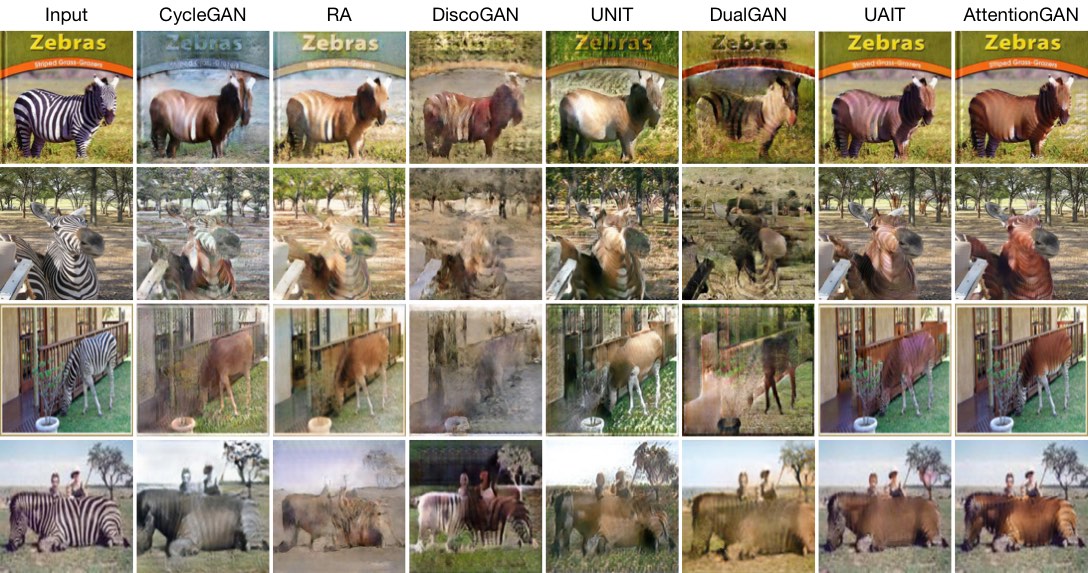}
	\caption{Different methods for mapping zebra to horse.}
	\label{fig:z2h}
	\vspace{-0.4cm}
\end{figure}

\begin{figure}[!t] \footnotesize
	\centering
	\includegraphics[width=1\linewidth]{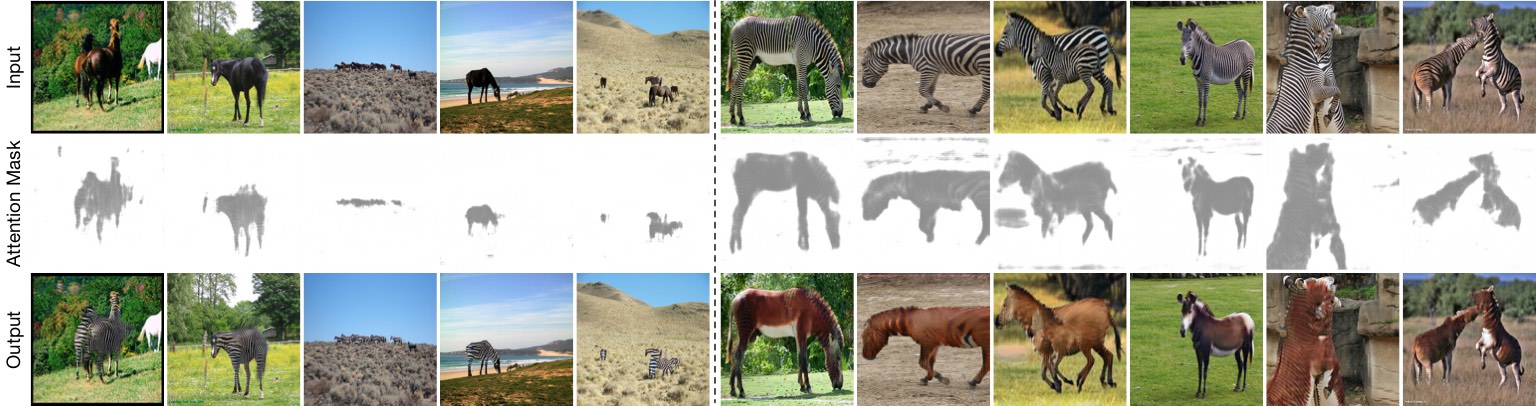}
	\caption{Attention masks on horse $\leftrightarrow$ zebra translation.}
	\label{fig:h2z_attention}
	\vspace{-0.4cm}
\end{figure}

We also compare the proposed AttentionGAN with GANimorph and CycleGAN in Fig.~\ref{fig:task}.
We see that the proposed AttentionGAN demonstrates a significant qualitative improvement over both methods.
Moreover, in the last row of Fig.~\ref{fig:task}, there is some blur in the bottom right corner of our result.
The reason is that our generator regards that location as foreground content and translates it to the target domain. This phenomenon rarely occurs in other samples (in Fig.~\ref{fig:h2z_attention}) because this sample is more difficult to learn and translation.

The results of zebra to horse translation are shown in Fig.~\ref{fig:z2h}.
We note that our method generates better results than all the baselines.
In summary, AttentionGAN is able to better alter the object of interest than existing methods by modeling attention masks in unpaired image-to-image translation tasks without changing the background at the same time.

\begin{figure}[!t] \footnotesize
	\centering
	\includegraphics[width=1\linewidth]{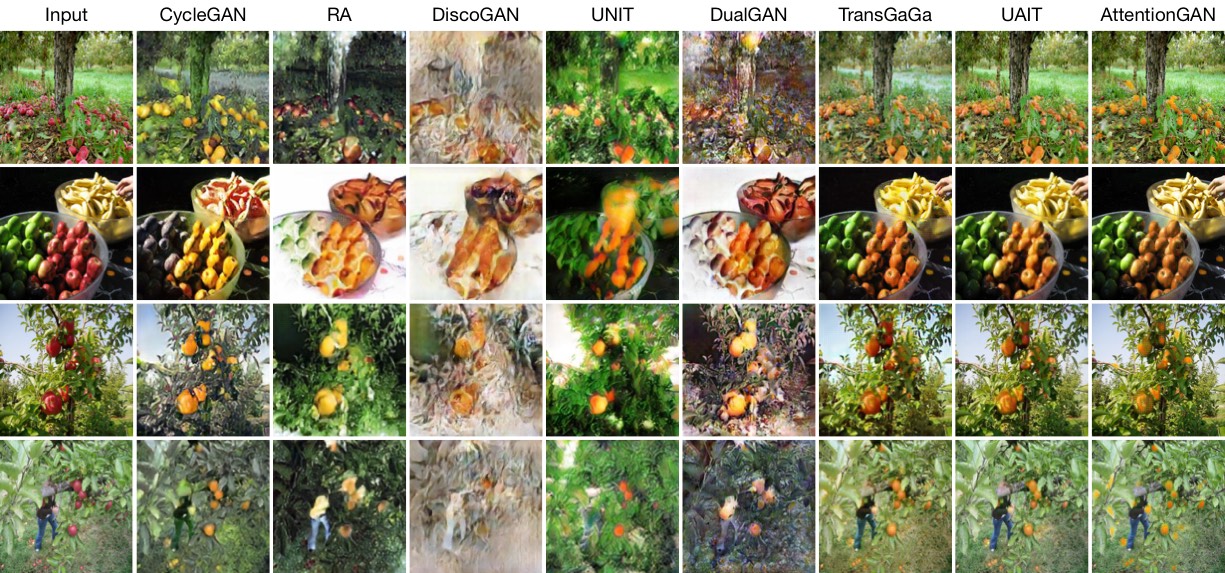}
	\caption{Different methods for mapping apple to orange.}
	\label{fig:a2o}
	\vspace{-0.4cm}
\end{figure}

\begin{figure}[!t] \footnotesize
	\centering
	\includegraphics[width=1\linewidth]{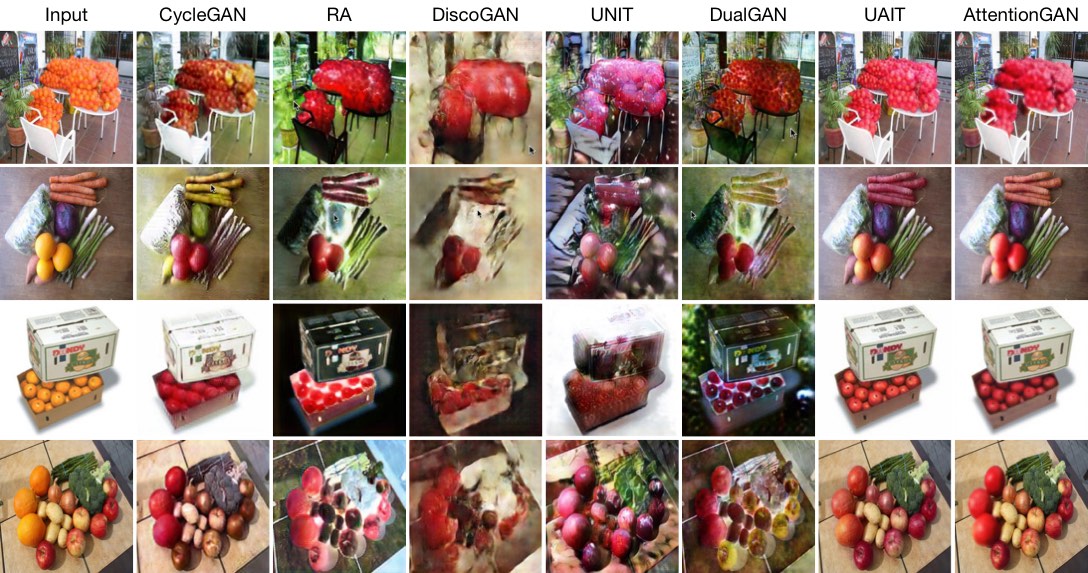}
	\caption{Different methods for mapping orange to apple.}
	\label{fig:o2a}
	\vspace{-0.4cm}
\end{figure}

\begin{figure}[!t] \footnotesize
	\centering
	\includegraphics[width=1\linewidth]{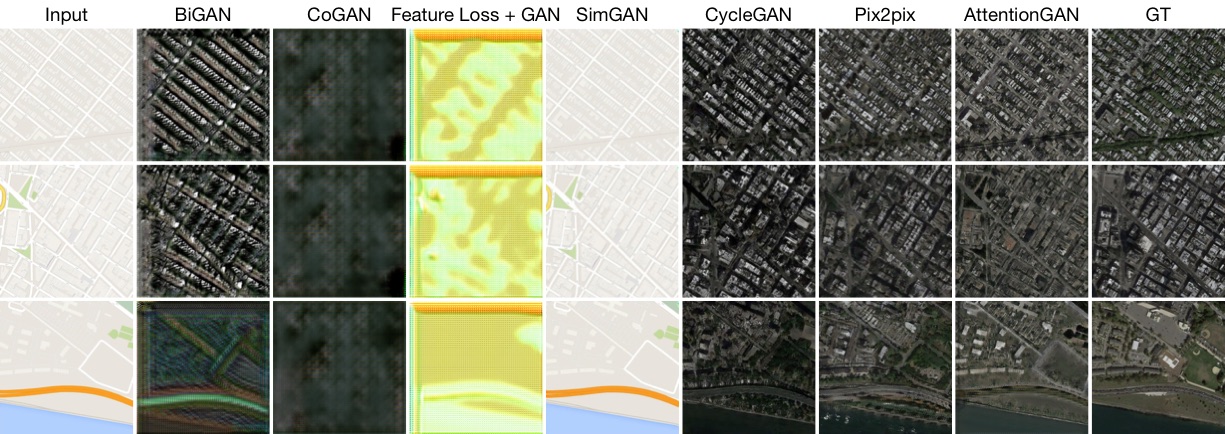}
	\caption{Different methods for mapping map to aerial photo.}
	\label{fig:l2m}
	\vspace{-0.4cm}
\end{figure}

\begin{figure}[!t] \footnotesize
	\centering
	\includegraphics[width=1\linewidth]{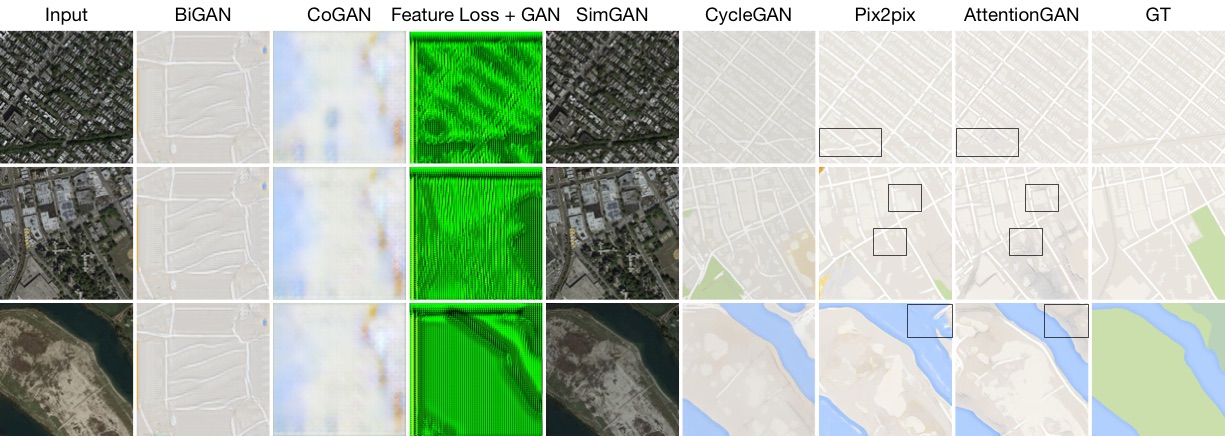}
	\caption{Different methods for mapping aerial photo to map.}
	\label{fig:m2l}
	\vspace{-0.4cm}
\end{figure}

\noindent \textbf{Results of Apple2Orange.}
The results compared with CycleGAN, RA, DiscoGAN, UNIT, DualGAN, TransGaGa, and UAIT are shown in Fig.~\ref{fig:a2o} and \ref{fig:o2a}.
We see that RA, DiscoGAN, UNIT, and DualGAN generate blurred results with lots of visual artifacts.
CycleGAN generates better results, however,  we can see that the background and other unwanted objects have also been changed, e.g., the banana in the second row of Fig.~\ref{fig:a2o}.
Both UAIT and our AttentionGAN can generate much better results than other baselines.
However, UAIT adds an attention network before each generator to achieve the translation of the relevant parts, which increases the number of network parameters.

\noindent \textbf{Results of Maps.}
Qualitative results of both translation directions compared with existing methods are shown in Fig.~\ref{fig:l2m} and \ref{fig:m2l}, respectively.
We note that BiGAN, CoGAN, SimGAN, Feature loss+GAN only generate blurred results with lots of visual artifacts.
The results generated by our method are better than those generated by CycleGAN.
Moreover, we compare AttentionGAN with the fully supervised Pix2pix, we see that AttentionGAN achieves comparable or even better results than Pix2pix as indicated in the black boxes in Fig.~\ref{fig:m2l}.

\begin{figure}[!t] \footnotesize
	\centering
	\includegraphics[width=0.85\linewidth]{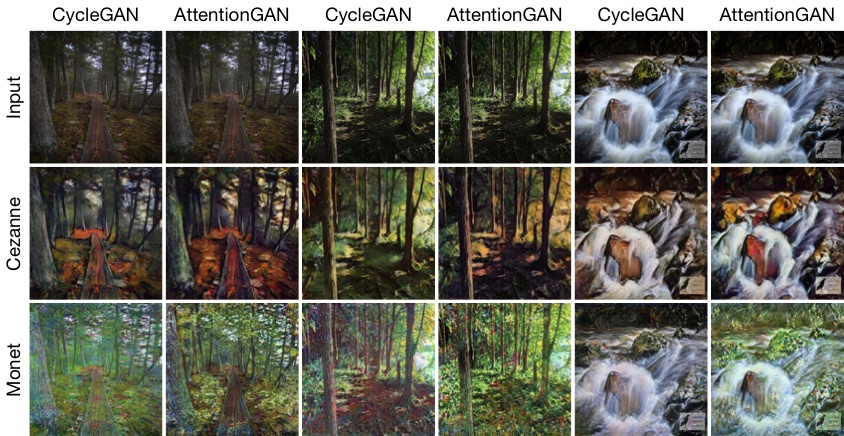}
	\caption{Different methods for style transfer.}
	\label{fig:style}
	\vspace{-0.4cm}
\end{figure}

\noindent \textbf{Results of Style Transfer.}
We also show the generation results of our AttentionGAN on the style transfer task. 
The results compared with the leading method, i.e., CycleGAN, are shown in Fig.~\ref{fig:style}.
We observe that the proposed AttentionGAN generates much sharper and diverse results than CycleGAN.

\begin{table}[!t] \small
	\centering
	\caption{KID $\times$ 100 $\pm$ std. $\times$ 100 for different methods. For this metric, lower is better. Abbreviations: (H)orse, (Z)ebra (A)pple, (O)range.}
	\resizebox{1\linewidth}{!}{
		\begin{tabular}{rcccc} 		    
			Method                                   & H $\rightarrow$ Z& Z $\rightarrow$ H     & A $\rightarrow$ O        & O $\rightarrow$ A \\ \midrule
			DiscoGAN~\cite{kim2017learning}        & 13.68 $\pm$ 0.28 & 16.60 $\pm$ 0.50      & 18.34 $\pm$ 0.75         & 21.56 $\pm$ 0.80 \\
			RA~\cite{wang2017residual}              & 10.16 $\pm$ 0.12 & 10.97 $\pm$ 0.26      & 12.75 $\pm$ 0.49         & 13.84 $\pm$ 0.78 \\
			DualGAN~\cite{yi2017dualgan}          & 10.38 $\pm$ 0.31 & 12.86 $\pm$ 0.50      & 13.04 $\pm$ 0.72         & 12.42 $\pm$ 0.88 \\
			UNIT~\cite{liu2017unsupervised}      & 11.22 $\pm$ 0.24 & 13.63 $\pm$ 0.34      & 11.68 $\pm$ 0.43         & 11.76 $\pm$ 0.51 \\
			CycleGAN~\cite{zhu2017unpaired}    & 10.25 $\pm$ 0.25 & 11.44 $\pm$ 0.38      & 8.48 $\pm$ 0.53          & 9.82 $\pm$ 0.51 \\
			UAIT~\cite{mejjati2018unsupervised}  & 6.93 $\pm$ 0.27  & 8.87 $\pm$ 0.26       & \textbf{6.44 $\pm$ 0.69} & 5.32 $\pm$ 0.48\\
			AttentionGAN                                  & \textbf{2.03 $\pm$ 0.64} & \textbf{6.48 $\pm$ 0.51}  & 10.03 $\pm$ 0.66 & \textbf{4.38 $\pm$ 0.42} \\
	\end{tabular}}
	\label{tab:kid}
	\vspace{-0.4cm}
\end{table}

\begin{table}[!t] \small
	\centering
	\caption{Preference score on both horse to zebra and apple to orange translation tasks. For this metric, higher is better. Abbreviations: (H)orse, (Z)ebra (A)pple, (O)range.}
		\begin{tabular}{rcccc} 		    
			Method                                    & H $\rightarrow$ Z  & Z $\rightarrow$ H  & A $\rightarrow$ O & O $\rightarrow$ A \\ \midrule
			UNIT~\cite{liu2017unsupervised}        & 1.83            & 3.57                        & 2.67                      & 2.91\\
			MUNIT~\cite{huang2018multimodal}   & 3.86            & 5.61                       &  6.23                      & 4.98\\
			DRIT~\cite{lee2018diverse}             & 1.27                 &  2.14                       & 1.09                       & 1.97\\
			CycleGAN~\cite{zhu2017unpaired}  & 22.12             &  21.65                     & 26.76                     & 26.14\\
			U-GAT-IT~\cite{kim2019ugatit}        & 33.17             &  31.39                    & 30.05                     & 29.89\\
			AttentionGAN                                & \textbf{37.75}  &  \textbf{35.64}        & \textbf{33.20}       & \textbf{34.11}\\
	\end{tabular}
	\label{tab:amt}
	\vspace{-0.4cm}
\end{table}

\begin{table}[!t] \small
	\centering
	\caption{The results of FID on the horse to zebra translation task. For this metric, lower is better.}
		\begin{tabular}{rc} 		    
			Method            & Horse to Zebra   \\ \midrule
			UNIT~\cite{liu2017unsupervised}        &  241.13                       \\
			CycleGAN~\cite{zhu2017unpaired}      &  109.36                     \\
			DA-GAN \cite{ma2018gan}      & 103.42  \\
			TransGaGa \cite{wu2019transgaga} & 95.81 \\
			SAT (Before Attention)~\cite{yang2019show}    & 98.90  \\
			SAT (After Attention)~\cite{yang2019show}    &  128.32  \\			
			AttentionGAN        & \textbf{68.55} \\
	\end{tabular}
	\label{tab:fid}
	\vspace{-0.4cm}
\end{table}

\begin{table}[!t] \small
	\centering
	\caption{AMT ``real vs fake'' results on maps $\leftrightarrow$ aerial photos. For this metric, higher is better.}
		\begin{tabular}{rcc} 		    
			Method                                                & Map to Photo               & Photo to Map \\ \midrule
			CoGAN~\cite{liu2016coupled}                     & 0.8 $\pm$ 0.7              &  1.3 $\pm$ 0.8 \\
			BiGAN/ALI~\cite{donahue2016adversarial,dumoulin2016adversarially}    & 3.2 $\pm$ 1.5              &   2.9 $\pm$ 1.2 \\
			SimGAN~\cite{shrivastava2017learning}                   & 0.4 $\pm$ 0.3             &   2.2 $\pm$ 0.7 \\
			Feature loss + GAN~\cite{shrivastava2017learning}  &  1.1 $\pm$ 0.8               &  0.5 $\pm$ 0.3 \\
			CycleGAN~\cite{zhu2017unpaired}                    & 27.9 $\pm$ 3.2              & 25.1 $\pm$ 2.9 \\
			Pix2pix~\cite{isola2016image}                    & 33.7 $\pm$ 2.6.              & 29.4 $\pm$ 3.2 \\
			AttentionGAN                                           & \textbf{35.18 $\pm$ 2.9}   & \textbf{32.4 $\pm$ 2.5} \\
	\end{tabular}
	\label{tab:m2l_amt}
	\vspace{-0.4cm}
\end{table}

\noindent \textbf{Quantitative Comparison.}
We follow UAIT~\cite{mejjati2018unsupervised} and adopt KID~\cite{binkowski2018demystifying} to evaluate the generated images by different methods.
The results of horse $\leftrightarrow$ zebra and apple $\leftrightarrow$ orange are shown in Table~\ref{tab:kid}.
We observe that AttentionGAN achieves the lowest KID on H $\rightarrow$ Z, Z $\rightarrow$ H and O $\rightarrow$ A translation tasks.
We note that both UAIT and CycleGAN produce a lower KID score on apple to orange translation (A $\rightarrow$ O) but have poor quality image generation as shown in Fig.~\ref{fig:a2o}.

Moreover, following U-GAT-IT~\cite{kim2019ugatit},  we conducted a perceptual study to evaluate the generated images on Horse2Zebra and Apple2Orange. 
Specifically, 50 participants are shown the generated images from different methods including AttentionGAN with source image, and asked to select the best generated image to the target domain.
The results are shown in Table~\ref{tab:amt}.
We observe that the proposed method outperforms other baselines including U-GAT-IT on these four tasks.

\begin{figure}[!t] \footnotesize
	\centering
	\includegraphics[width=1\linewidth]{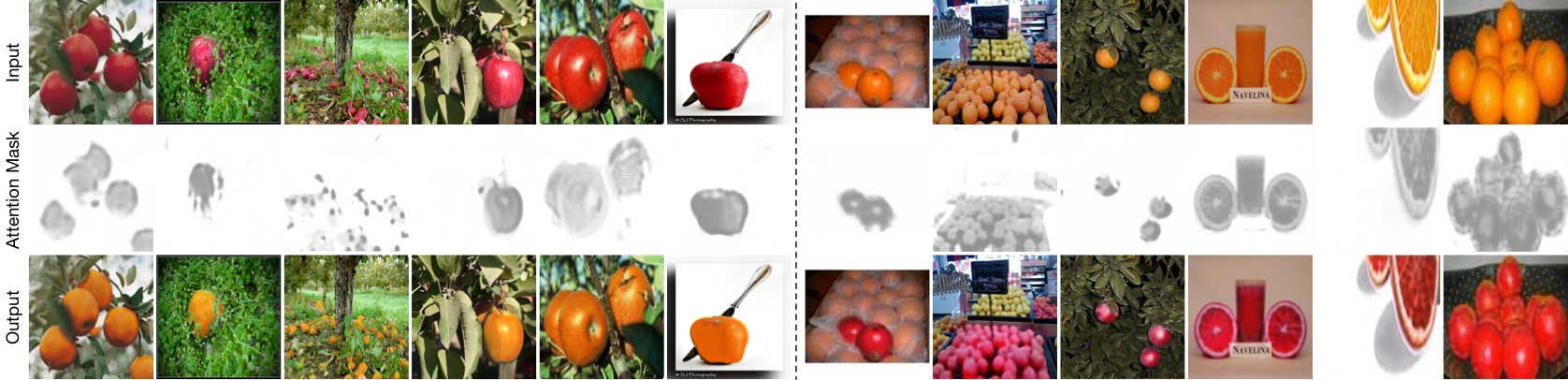}
	\caption{Attention masks on apple $\leftrightarrow$ orange translation.}
	\label{fig:o2a_attention}
	\vspace{-0.4cm}
\end{figure}

\begin{figure}[!t] \footnotesize
	\centering
	\includegraphics[width=0.85\linewidth]{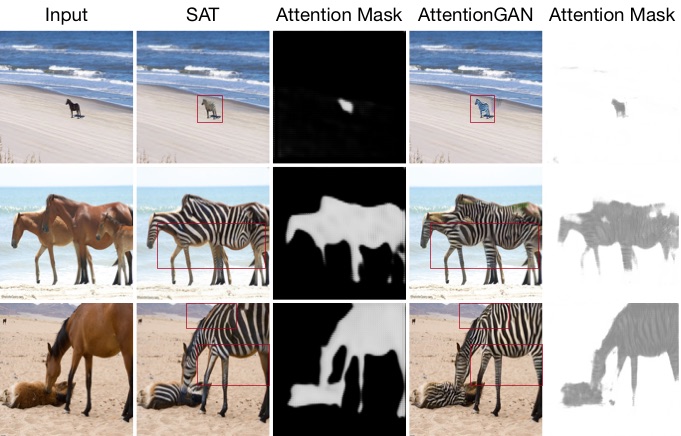}
	\caption{Attention masks compared with SAT~\cite{yang2019show} on horse to zebra translation.
	}
	\label{fig:attention_comparsion}
	\vspace{-0.4cm}
\end{figure}

Next, we follow SAT~\cite{yang2019show} and adopt FID~\cite{heusel2017gans} to measure the distance between generated samples and target samples. 
We compute FID for horse to zebra translation and the results compared with SAT, CycleGAN, DA-GAN, TransGaGa, and UNIT are shown in Table~\ref{tab:fid}. 
We observe that the proposed model achieves significantly better FID than all baselines. 
We note that SAT with attention has worse FID than SAT without attention, which means using attention might have
a negative effect on FID because  there might be some correlations between foreground and background in the target
domain when computing FID.
While we did not observe such a negative effect on AttentionGAN.
Qualitative comparison with SAT is shown in Fig.~\ref{fig:attention_comparsion}.
We observe that the proposed AttentionGAN achieves better results than SAT. 

Finally, we follow CycleGAN and adopt AMT score to evaluate the generated images on the  map $\leftrightarrow$ aerial photo translation task.
Participants were shown a sequence of pairs of images, one real image and one fake generated by our method or exiting methods, and asked to click on the image they thought was real. 
Comparison results of both translation directions are shown in Table~\ref{tab:m2l_amt}.
We observe that the proposed AttentionGAN generate the best results compared with the leading methods and can fool participants on around 1/3 of trials in both translation directions.

\begin{figure}[!t] \footnotesize
	\centering
	\includegraphics[width=1\linewidth]{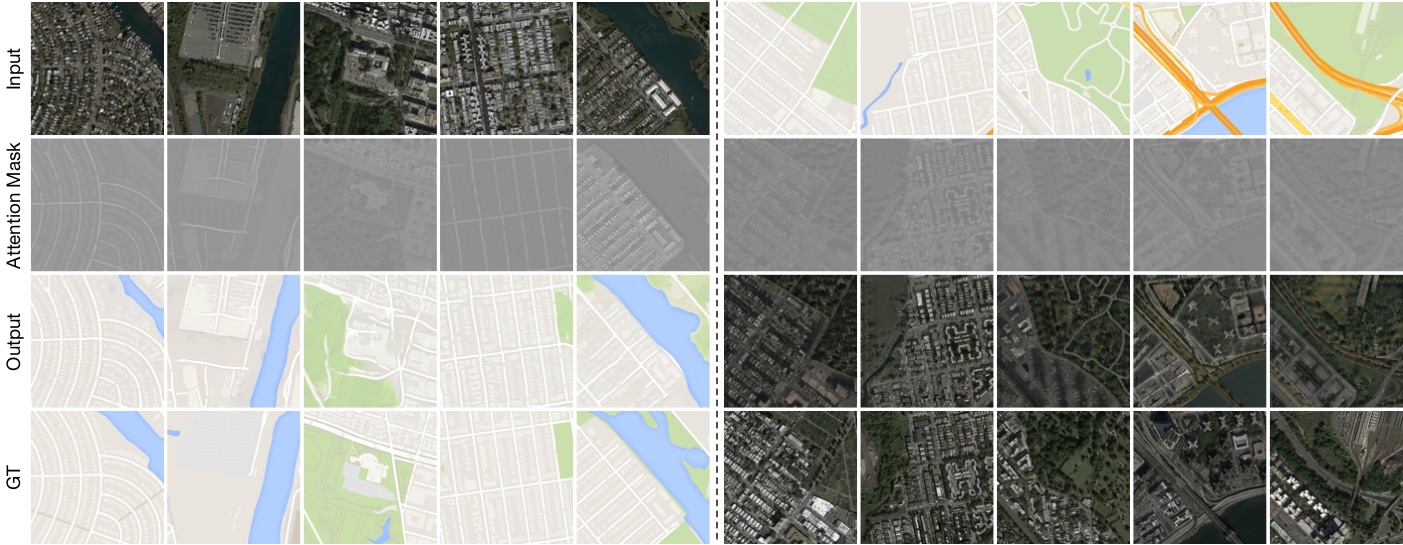}
	\caption{Attention masks on aerial photo $\leftrightarrow$ map translation.
	}
	\label{fig:l2m_attention}
	\vspace{-0.4cm}
\end{figure}

\noindent \textbf{Visualization of Learned Attention Masks.}
The results of both horse $\leftrightarrow$ zebra and apple $\leftrightarrow$ orange translations are shown in Fig.~\ref{fig:h2z_attention} and \ref{fig:o2a_attention}, respectively.
We see that our AttentionGAN is able to learn relevant image regions and ignore the background and other irrelevant objects.
Moreover, we also compare with the most recently method, SAT~\cite{yang2019show}, on the learned attention masks.
Results are shown in Fig.~\ref{fig:attention_comparsion}.
We observe that the attention masks learned by our method are much more accurate than those generated by SAT, especially in the boundary of attended objects.
Thus our method generates a more photo-realistic object boundary than SAT in the translated images, as indicated in the red boxes in Fig.~\ref{fig:attention_comparsion}.

The results of map $\leftrightarrow$ aerial photo translation are shown in Fig.~\ref{fig:l2m_attention}.
Note that although images of the source and target domains differ greatly in the appearance, the images of both domains are structurally identical.
Thus the learned attention masks highlight the shared layout and structure of both source and target domains. 
Thus, we can conclude that AttentionGAN can handle both images requiring large shape changes and images requiring holistic changes.

%% file: 5conclusions.tex
\section{Conclusion}
\label{sec:con}
We propose a novel AttentionGAN for both unpaired image-to-image translation and multi-domain image-to-image translation tasks. 
The generators in AttentionGAN have the built-in attention mechanism, which can preserve the background of the input images and discovery the most discriminative content between the source and target domains by producing attention masks and content masks.
Then the attention masks, content masks, and the input images are combined to generate the target images with high quality.
Extensive experimental results on several challenging tasks demonstrate that the proposed AttentionGAN can generate better results with more convincing details than numerous state-of-the-art methods.

\section*{Acknowledgment}
This work is partially supported by National Natural Science Foundation of China (No.62073004), Shenzhen Fundamental Research Program (No.GXWD20201231165807007-20200807164903001), by the EU H2020 AI4Media No. 951911 project and the Italy-China collaboration project TALENT:2018YFE0118400.